\documentclass[letterpaper]{article} 
\usepackage{aaai2026}  
\usepackage{times}  
\usepackage{helvet}  
\usepackage{courier}  
\usepackage[hyphens]{url}  
\usepackage{graphicx} 
\usepackage{natbib}  
\usepackage{caption} 
\usepackage{tcolorbox}
\usepackage{algorithm}
\usepackage{algorithmic}
\usepackage{booktabs}
\usepackage{graphicx}
\usepackage{multirow}
\usepackage{newfloat}
\usepackage{listings}
\DeclareCaptionStyle{ruled}{labelfont=normalfont,labelsep=colon,strut=off} 
\floatstyle{ruled}
\newfloat{listing}{tb}{lst}{}
\floatname{listing}{Listing}
\usepackage{colortbl}
\definecolor{mycolor}{rgb}{0.925, 0.945, 0.980}
\definecolor{mygray}{rgb}{0.886, 0.886, 0.886}
\definecolor{mypink}{rgb}{0.992,0.921,0.929}
\definecolor{mygreen}{rgb}{0.901,0.950,0.937}
\usepackage{algorithm}
\usepackage{algorithmic}
\usepackage{booktabs}
\usepackage{graphicx}
\usepackage{mathrsfs}
\usepackage{newfloat}
\usepackage{listings}
\usepackage[table]{xcolor} 
\usepackage{amsthm,amsmath,amssymb}

\title{History-Aware Reasoning for GUI Agents}
\author {
    Ziwei Wang\textsuperscript{\rm 1,2},
    Leyang Yang\textsuperscript{\rm 1,2},
    Xiaoxuan Tang\textsuperscript{\rm 3},
    Sheng Zhou \textsuperscript{\rm 1}\thanks{Corresponding Author},\\
    Dajun Chen \textsuperscript{\rm 3},
    Wei Jiang \textsuperscript{\rm 3},
    Yong Li \textsuperscript{\rm 3}\footnotemark[1]
    }
\affiliations {
    \textsuperscript{\rm 1}Zhejiang Key Laboratory of Accessible Perception and Intelligent Systems, Zhejiang University\\
    \textsuperscript{\rm 2}College of Computer Science and Technology, Zhejiang University\\
    \textsuperscript{\rm 3}Ant Group\\
    \{wangziwei98, yangleyang, zhousheng\_zju\}@zju.edu.cn\\
    \{tangxiaoxuan.txx, chendajun.cdj, liyong.liy, jonny.jw\}@antgroup.com
}

\usepackage{bibentry}

\newcommand{\model}{HAR-GUI-3B}

\begin{document}

\maketitle

\begin{abstract}
Advances in Multimodal Large Language Models have significantly enhanced Graphical User Interface (GUI) automation.
Equipping GUI agents with reliable episodic reasoning capabilities is essential for bridging the gap between users’ concise task descriptions and the complexities of real-world execution. 
Current methods integrate Reinforcement Learning (RL) with System-2 Chain-of-Thought, yielding notable gains in reasoning enhancement.
For long-horizon GUI tasks, historical interactions connect each screen to the goal-oriented episode chain, and effectively leveraging these clues is crucial for the current decision.
However, existing native GUI agents exhibit weak short-term memory in their explicit reasoning, interpreting the chained interactions as discrete screen understanding, i.e., unawareness of the historical interactions within the episode. 
This history-agnostic reasoning challenges their performance in GUI automation.
To alleviate this weakness, we propose a History-Aware Reasoning (HAR) framework, which encourages an agent to reflect on its own errors and acquire episodic reasoning knowledge from them via tailored strategies that enhance short-term memory in long-horizon interaction.
The framework mainly comprises constructing a reflective learning scenario, synthesizing tailored correction guidelines, and designing a hybrid RL reward function. 
Using the HAR framework, we develop a native end-to-end model, \model, which alters the inherent reasoning mode from history-agnostic to history-aware, equipping the GUI agent with stable short-term memory and reliable perception of screen details. Comprehensive evaluations across a range of GUI-related benchmarks demonstrate the effectiveness and generalization of our method.

\end{abstract}

\begin{links}
    \link{Code}{https://github.com/BigTaige/HAR-GUI}
\end{links}

\section{Introduction}
\label{sec:intro}
Graphical User Interface (GUI) agents have witnessed remarkable advancements with the integration of advanced Multimodal Large Language Models (MLLMs), enabling autonomous manipulation of end-user devices via tailored development for GUI scenarios~\cite{qin2025ui}. 
Such capability holds significant value in applications like accessibility and automated testing. Early methods rely on MLLMs with generic multimodal understanding capability, using function calls and context engineering to manually construct workflows for GUI automation~\cite{mobileagentv2,chen2024automanual}. However, these methods demanded meticulous prompting design and faced performance bottlenecks due to models’ sensitivity to instructions~\cite{zhuo-etal-2024-prosa}. Reliance on expert experience and handcrafted instructions limits scalability across GUI-oriented tasks in device ecosystems, and the closed-source nature of these large-scale MLLMs further restricts domain-specific optimization, underscoring the need for more adaptable solutions.

In contrast to screen understanding tasks~\cite{widgetcaption,screenQA,tap,chen2021websrc}, GUI automation is far more challenging, a GUI agent must integrate reliable reasoning with the ability to interpret a concise task description and then interact with the device step-by-step~\cite{survey2}. 
In long-horizon GUI tasks, historical interaction information connects each screen to the entire episode chain shaped by the user’s overall goal, and the agent’s ability to reliably perceive this information is crucial for action decision at the current screen status.
By integrating GUI-specific knowledge, current MLLMs~\cite{bai2025qwen2.5,mimo-vl} can adapt to GUI scenarios, and post-training these domain-specific foundation models for GUI agent development has produced promising results~\cite{wang2025mp}, particularly when adopting a System-2 reasoning mode~\cite{qin2025ui}, in which GUI agents perform explicit logical reasoning before predicting actions, and applying RL~\cite{shao2024deepseekmath} to further enhance their reasoning capability.
However, despite the remarkable gains in screen perception~\cite{gui-g1} and reasoning enhancement achieved by existing methods~\cite{luo2025gui}, we find that current native GUI agents exhibit a notable short-term memory weakness during episodic reasoning. Specifically, \textbf{their reasoning mode is agnostic to historical interactions, degrading chained interactions to discrete screen understanding}. This "history-agnostic" stems from the foundation MLLM's inherent Chain-of-Thought (CoT), undermining performance in episodic reasoning that need to leverage the previous execution clues.

Current advanced methods focus on enhancing the overall reasoning of GUI agents, whereas exploration of specific defects in episodic reasoning remains limited. Drawing on heuristic learning, several methods leverage hard samples, i.e., instances that are previously incorrectly inferred, and apply RL for reflective training, encouraging the agent to acquire unfamiliar domain-specific reasoning knowledge from these samples~\cite{qin2025ui,infigui-r1}. RL primarily steers the model toward reward-bearing trajectories, thereby refining its output strategy from generating multiple candidates (pass@k) to a single, high-confidence solution (pass@1). However, this process remains fundamentally limited by the foundation model's prior knowledge.~\cite{yue2025doesreinforcementlearningreally}. Without introducing external GUI-specific reasoning knowledge, the agent’s short-term memory weakness persists. Moreover, these methods perform RL with inference-format instructions~\cite{lu2025ui,wu2025gui,reachagent}, i.e., training instructions identical to the inference phase, yet our experiments indicate that the optimization signal primarily affects action-level prediction without enhancing the agent’s inherent reasoning mode.

In this work, we aim to enhance the reasoning capabilities of the GUI agent by equipping it with short-term memory for episodic reasoning, enabling explicit historical-context awareness in its System-2 CoT, which in turn fortifies its overall performance on GUI-oriented tasks.
Consequently, we propose a \underline{\textbf{H}istory-\textbf{A}ware \textbf{R}easoning \textbf{(HAR)} framework} for reflective training with tailored GUI reasoning enhancement strategies. Our method comprises three key components: \textbf{(\textit{i})} \textbf{Reflective learning scenario construction}; \textbf{(\textit{ii})} \textbf{Tailored correction guidelines synthesis}; \textbf{(\textit{iii})} \textbf{A hybrid RL reward function to encourage historical awareness in the GUI agent}.
Using the HAR framework, we alter the reasoning mode of the GUI agent from being history-agnostic to being history-aware. This allows for explicit integration of historical context and cognitive correction via error-centric self-evolution. As a result, the GUI agent emerges stable short-term memory, allowing it to flexibly perceive the episode’s chained historical clues and make reasonable use of it. This strengthened reasoning enables the GUI agent to handle long-horizon interactions and achieving consistent and persistent gains across GUI-oriented tasks.

We evaluate our method on diverse, widely used GUI-related benchmarks and manually annotate a challenging Chinese mini-program benchmark for out-of-distribution (OOD) generalization comparison. Experiments demonstrate that our method outperforms current advanced methods with similar parameters on multiple GUI-related benchmarks and rivals larger models in OOD scenarios. Our main contributions are as follows:

\begin{itemize}
    \item We propose HAR, a framework that employs tailored strategies and reflective learning to deepen the agent’s GUI-specific knowledge and transform its reasoning from history-agnostic to history-aware, enhancing its short-term memory for episodic reasoning.
    \item Using the HAR framework, we develop \textbf{\model}, a GUI-tailored native model with reliable episodic reasoning and screen perception, and demonstrate its effectiveness across a range of GUI-related benchmarks.
    \item \model\ shows consistent generalization. In OOD scenarios, it outperforms SOTA methods with similar parameter sizes and competes with much larger models.
\end{itemize}

\section{Related Works}
\label{sec:related_work}
Early methods rely on sophisticated context engineering and MLLM function calls to construct workflows for GUI automation, \textit{e.g.}, ReAct~\cite{react}, Reflexion~\cite{shinn2023reflexion}, Expel~\cite{zhao2024expel} and AppAgent~\cite{zhang2025appagent}. While these methods show promise, their effectiveness relies on the MLLM's general capability, manual experience, and sensitivity to instructions.
Subsequently, researchers enhance MLLMs' GUI perception by training small-parameter models with domain-specific data, yielding native GUI agents such as SeeClick~\cite{cheng2024seeclick}, MP-GUI~\cite{wang2025mp} and ShowUI~\cite{lin2025showui} via supervised fine-tuning (SFT). With RL's success in reasoning enhancement~\cite{shao2024deepseekmath}, several methods employ RL to train agents~\cite{bai2024digirl}, achieving remarkable reasoning and generalization capabilities. Domain-specific pre-training further enables MLLMs to tackle complex GUI tasks using concise instructions~\cite{bai2025qwen2.5}.
The slow thinking mode of System-2 reasoning emerges as an effective approach for enhancing GUI agents (\textit{e.g.}, UI-TARS~\cite{qin2025ui}, InfiGUI-R1-3B~\cite{infigui-r1}, GUI-R1-3B~\cite{luo2025gui} and UI-R1-3B~\cite{lu2025ui}). Meanwhile, several methods steer the agent's attention to specific content via instructional constraints~\cite{mobileagentv2,zhang2025appagent}. However, this manual prompting approach contradicts the natural reasoning behavior of the foundation models on GUI automation, resulting in unstable performance and hallucinations. Alternatively, we construct a reflective learning scenario and tailored strategies to equip the agent with competent GUI reasoning capability. 

\begin{figure*}[ht]
    \centering
    \includegraphics[width=0.9\linewidth]{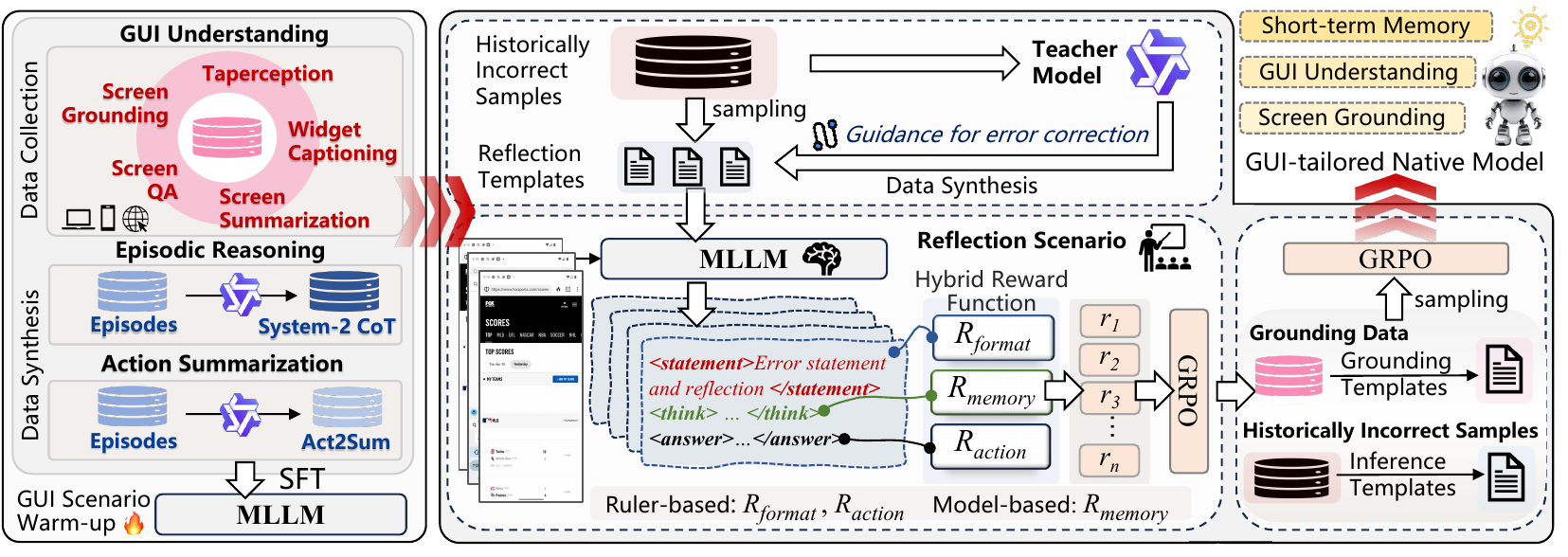}
    \caption{Overview of the Histoty-Aware Reasoning (HAR) framework. HAR framework is an error-centric learning approach designed to enhance the reasoning capability of the GUI agent by performing error-aware cognitive correction within a tailored reflection scenario. The framework consists of two critical training stages: \textbf{(\textit{i})} \textbf{GUI Scenario Warm-up Stage.} During this phase, comprehensive domain-specific knowledge is injected into the agent via GUI-related data collection and synthesis for knowledge distillation. \textbf{(\textit{ii})} \textbf{Learning From Failure Stage.} In this stage, the agent's short-term memory is enhanced. It involves a round of RL within the reflection scenario to perform error-aware cognitive corrections that boost episodic reasoning, followed by another round of RL employing a task-mixing training strategy to assist the GUI agent perceive screen visual details.}
    \label{fig:IFF_main_framework}
\end{figure*}

\section{HAR Framework} \label{sec:iff-freamwork}
An overview of the Histoty-Aware Reasoning (HAR) framework is illustrated in Fig.\ref{fig:IFF_main_framework}. In this section, we first define the GUI episode reasoning task and then introduce the critical training stages (\textit{GUI scenario warm-up} and \textit{learning from failure}) in the HAR framework and the details of the proposed methods in each training stage.
\subsection{Task Definition}
We first formulate the execution of the GUI agent, let $\mathcal{D}$ denotes an episode chain with an overall goal $\mathcal{G}$, $\mathcal{D}$ = $(\mathcal{G},(O_1,\mathcal{A}_1),\dots,(O_n,\mathcal{A}_n))$, and observation $O_t$=$(I_t, \mathbb{P})$, where $\mathcal{A}_t$ $\in$ $\mathbb{A}$ means the action executed by the agent at time step $t \in [1,n]$, $\mathbb{A}$ is the pre-defined action space (\textit{e.g.}, {\tt CLICK}, {\tt SCROLL} and {\tt TYPE}), $I_t$ is the screen image, $\mathbb{P}$ denotes the textual instruction template. The task can be formulated as a Markov Decision Process $P(\mathcal{A}_t|O_{\leq t},A_{<t},\mathcal{G})$. Since step-by-step execution, chained interaction histories $\mathscr{T}_{<t}$ are critical for the current decision. While some methods concatenate $(I_{<t},\mathcal{A}_{<t})$ pairs into the input for $\mathscr{T}_{<t}$ transmission, computing image tokens is expensive, especially for high-resolution devices~\cite{lin2025showui}. To balance computational costs and performance, most research conveys $\mathscr{T}_{<t}$ via textual modality~\cite{infigui-r1,luo2025gui} by summarizing interactions and integrating them into $\mathbb{P}$, which is also a strategy adopted in this work.

\subsection{GUI Scenario Warm-up}
Due to the weak screen perception of the foundation model $\mathcal{M}_{\text{\textit{base}}}$ in handling GUI-oriented tasks, in this training stage, we collect and synthesize comprehensive GUI-related data for the injection of domain-specific knowledge through supervised fine-tuning (SFT).

\noindent\textbf{GUI Understanding Enhancement.}
Effective screen perception is crucial for GUI agents to conduct GUI-oriented tasks~\cite{cheng2024seeclick,wang2025mp}, \textit{e.g.}, screen analysis~\cite{widgetcaption,screen2words} and screen question-answering~\cite{screenai}. To enhance the fundamental screen understanding of $\mathcal{M}_{\text{\textit{base}}}$, we collect comprehensive domain-specific tasks from public sources, including graphic caption generation, UI-element clickability prediction, screen question answering, screen summary, and screen grounding\footnote{For grounding task, we use the data format of text2point, and we also experiment with text2box and a mixture of the two formats. Experiments show that text2point performs best in our framework.}. These data cover platforms with multiple resolutions, including mobile, desktop, and website.

\noindent\textbf{Action Summarization Integration.} The GUI agent is required to interact with the screen for multiple rounds, analyze the current screen status, and interpret the overall goal to make the ongoing decision. Thus, the agent needs to be equipped with a reliable action understanding capability, i.e., it is crucial for the agent to effectively comprehend the semantics of the action at the episode level. 

In this part, we introduce an action summarization task designed to enhance the action semantic understanding of $\mathcal{M}_{\text{\textit{base}}}$. Specifically, we construct instruction templates that prompt a teacher model to synthesize \textbf{Action-to-Summary (Act2Sum)} data. Specifically, let $\mathcal{G}$ denote the goal, $\mathcal{I}_t$ the current screen image, $\mathcal{A}_t$ the corresponding action, $\mathbb{P}_{\text{\textit{sum}}}$ the instruction template, and $\mathcal{M}_{\text{\textit{teacher}}}(\cdot,\theta)$ the teacher model with parameters $\theta$. The process can be expressed as follows,
\begin{equation}
    \label{eq:act2sum}
    O^t_{Act2Sum} = \mathcal{M}_{\text{\textit{teacher}}}(\mathcal{I}_t,\mathbb{P}_{\text{\textit{sum}}}(\mathcal{G},\mathcal{A}_t),\theta)
\end{equation}

where $O^t_{Act2Sum}$ is the action summary corresponding to $\mathcal{A}_t$, $\mathcal{I}_t$, and $\mathcal{G}$.
We construct the Act2Sum dataset by pairing each summary $O^t_{Act2Sum}$ with its input $(\mathcal{I}_t,\mathbb{P}_{sum}(\mathcal{G},\mathcal{A}_t))$.

Since the input explicitly contains the episode’s goal, the synthesized content is goal-oriented and semantically rich, which assists $\mathcal{M}_{\text{\textit{base}}}$ in generating globally aware historical interaction records during episodic reasoning.\\
\noindent\textbf{System-2 Reasoning Data Distillation.} Compared with System-1 reasoning paradigm (\textit{fast thinking mode that generates answers directly}), explicitly generate intermediate thought processes via CoT before answering, System-2 reasoning (\textit{slow thinking}) denotes intentional, organized, and analytical thought, improving agents to tackle complex and multi-step tasks~\cite{qin2025ui}. To enhance the reasoning capability of $\mathcal{M}_{\text{\textit{base}}}$, we utilize the reasoning advanced $\mathcal{M}_{\text{\textit{teacher}}}$ to synthesize System-2 CoT for each instance of episodes. Specifically, we build the instruction template (\textit{inference format}) and conduct the inference with $\mathcal{M}_{\text{\textit{teacher}}}$. Afterwards, we filter out positive samples through evaluation methods and collect these synthetic samples with System-2 CoT as training data. Via training, we achieve high-quality GUI reasoning knowledge injection from $\mathcal{M}_{\text{\textit{teacher}}}$ to $\mathcal{M}_{\text{\textit{base}}}$.

\subsection{Learning From Failure} \label{sec:2round_RL}
After the warm-up stage, we yield a GUI-enhanced agent $\mathcal{M}_{\text{\textit{warm-up}}}$ from $\mathcal{M}_{\text{\textit{base}}}$. Inspired by the heuristic principle that \textit{"Adversity is the crucible of growth"}, we construct a reflection scenario for training to encourage $\mathcal{M}_{\text{\textit{warm-up}}}$ to grasp domain-sepcific reasoning knowledge from hard samples via self-evolution, thereby improving the reasoning of GUI agent when facing complex GUI-oriented tasks. We first perform inference using $\mathcal{M}_{\text{\textit{warm-up}}}$ on our episode data and flag error instances as historical incorrect samples $\mathbb{D}_{his}$.

\noindent\textbf{Guidance Synthesis for Error Correction.}
Currently, there are few methods conduct RL for reflective training on hard samples~\cite{wu2025gui,liu2025infiguiagent}. However, these methods are trained under an inference-style pattern. Although RL can narrow the output space from pass@k to pass@1 to improve the success rate through exploration~\cite{yue2025doesreinforcementlearningreally}, enhancing short-term memory remains a challenge. This requires altering the GUI agent's entrenched reasoning CoT established during pre-training.

Inspired by curriculum learning~\cite{liu2024let}, we conduct error analysis on samples in $\mathbb{D}_{his}$ by prompting $\mathcal{M}_{\text{\textit{teacher}}}$ to generate no more than three error correction guidelines $\mathbb{G}$ for each incorrect instance. Thus, $\mathbb{G}$ can serve as external GUI reasoning knowledge to facilitate $\mathcal{M}_{\text{\textit{warm-up}}}$ in clue-oriented reasoning enhancement and hard samples correction. We can formulate this process as follows,
\begin{equation}
    \label{eq:guidance}
    \begin{aligned}
&\mathbb{G}_t = \mathcal{M}_{\text{\textit{teacher}}}(\mathcal{I}_t,\mathbb{P}_{\text{\textit{guidance}}}(\hat{x}_t),\theta),\\
     &\hat{x}_t=(\mathcal{G}, \mathscr{T}_{<t}, \mathcal{A}_t, \mathcal{A}_t^{error},\mathcal{C}_t^{error})\in \mathbb{D}_{his}
\end{aligned}    
\end{equation}

where $\mathbb{G}_t$ means the tailored guidelines corresponding to $x_t$, $\mathcal{A}_t$ is the ground truth, $\mathbb{P}_{\text{\textit{guidance}}}(\cdot)$ represents the instruction, $\mathcal{A}_t^{error}$ denotes the incorrect action conducted by $\mathcal{M}_{\text{\textit{warm-up}}}$ and $\mathcal{C}_t^{error}$ is the CoT corresponding to $\mathcal{A}_t^{error}$.

\noindent\textbf{Reflection Scenario Construction.} 
Inspired by earlier methods that employ MLLM function calls and handcrafted, prompt-based reflection to correct erroneous execution paths~\cite{react,mobileagentv2}, current approaches adopt inference-format instructions and perform RL training on hard samples to enhance GUI agents~\cite{wu2025gui,liu2025infiguiagent}. 

As discussed, the weak domain-specific knowledge of foundation models in GUI automation results in a key observation: RL-based exploration under the inference-format instruction struggles both to correct hard samples and to enhance the short-term memory of GUI agents. Thus, we construct a reflection scenario for cognitive correction and introduce external GUI reasoning knowledge tailored for each instance in $\mathbb{D}_{his}$, i.e., guidelines $\mathbb{G}$, to assist the GUI agent in its self-evolution during clue-guided exploration. The process can be formulated as,
\begin{equation}
    \label{eq:reflection}
    \begin{aligned}
    &\mathcal{O}_t^* = \mathcal{M}_{\text{\textit{warm-up}}}(\mathcal{I}_t,\mathbb{P}_{\text{\textit{reflection}}}(x_t),\theta),\\
    &x_t=(\mathcal{G},\mathbb{G}_t,\mathscr{T}_{<t},\mathcal{A}_t^{error},\mathcal{C}_t^{error})\in\mathbb{R}_{his}
    \end{aligned}
\end{equation}

where $\mathcal{A}_t^*,\mathcal{C}_t^*,\mathcal{S}_t^*=\mathcal{O}_t^*$, $\mathbb{P}_{\text{\textit{reflection}}}$ denotes the instruction of the reflection format, $\mathcal{A}_t^*$ represents the action predicted by $\mathcal{M}_{\text{\textit{warm-up}}}$, $\mathcal{C}_t^*$ and $\mathcal{S}_t^*$ mean the corresponding CoT, as well as the statement and reflection on the previous incorrect prediction\footnote{These items will be extracted based on predefined tags in $\mathcal{O}_t^*$.}, respectively.

\noindent\textbf{Error-Aware Cognitive Correction.} Since we aim to guide the GUI agent to grasp episodic reasoning knowledge through self-exploration without supervisory signal constraints, we identify that the GRPO algorithm~\cite{shao2024deepseekmath} aligns well with this goal. Specifically, GRPO first generates $N$ candidate responses $\{\mathcal{O}_{t,i}^*\}_{i=1:N}$ for a task query. Each response $\mathcal{O}_{t,i}^*$ yields a reward $r_i$ via action execution. Then, normalizing rewards to compute the relative advantage ${E}_i$ of each response: 
\begin{equation}
    {E}_i = \frac{r_i - {mean}(\{r_1, \dots, r_N\})}{{std}(\{r_1, \dots, r_N\})}
\end{equation}

where $mean(\cdot)$ and $std(\cdot)$ denote the reward distribution’s mean and standard deviation. In our reflection scenario, each response $\mathcal{O}_{t,i}^*$ has three components: $\mathcal{A}_{t,i}^*$, $\mathcal{C}_{t,i}^*$, and $\mathcal{S}_{t,i}^*$. To make fuller utilize of the output signals, each sample's reward $r_i$ consists of multiple parts. First, a rule-based format reward $r_i^{\text{\textit{format}}}$ is used to check whether $\mathcal{O}_{t,i}^*$ conforms to the output format required by instruction $\mathbb{P}_{\text{\textit{reflection}}}$. If it matches, $r_i^{\text{\textit{format}}}$ = 1, otherwise it is $0$. Next, we define the action reward $r_i^{action}$. If the predicted action $\mathcal{A}_{t,i}^*$ matches the label, the reward is $1$, otherwise it is $0$. 

Note that in real-world execution, actions involving screen coordinates are frequently called, and our experiments found that the correct execution of such actions (\textit{e.g.}, {\tt CLICK}) is challenging for GUI agents. Thus, for this type of actions, we assign a higher reward to enhance the GUI agent’s perception of screen details. Specifically, let $P$ = $(p_x,p_y)$ and $P^*$ = $(p_x^*,p_y^*)$ represent the absolute coordinates $(x,y)$ of the label and prediction, respectively. Then, we calculate the scalar action reward $r_i^{action}$ as follows,
\begin{equation}
\label{eq:click_reward_1}
\begin{aligned}
    &\mathcal{F}_{\text{\textit{dist}}}(P_1,P_2) = \sqrt{(x_1-x_2)^2+(y_1-y_2)^2},\\
    &  P_1=(x_1,y_1), P_2=(x_2,y_2)\\
\end{aligned}
\end{equation}
\begin{equation}
\label{eq:click_reward_2}
\begin{aligned}
    &\mathcal{F}_{\text{\textit{abs}}}(P_1,P_2,\tau) = 
    \left\{ 
    \begin{aligned} 
    &1-\mathcal{F}_{\text{\textit{dist}}}(P_1,P_2)/\tau, \ \text{if} \ \mathcal{F}_{\text{\textit{dist}}}< \tau \\
    & 0, \text{else}
    \end{aligned}\right.\\
\end{aligned}
\end{equation}
\begin{equation}
\label{eq:click_reward}
\begin{aligned}
    & P_{norm}=(p_x / w,p_y/h),P_{norm}^*=(p_x^*/w,p_y^*/h)\\
&\mathcal{D}_{\text{\textit{norm}}}=\mathcal{F}_{\text{\textit{dist}}}(P_{norm},P_{norm}^*)\\
    &r_i^{action} = \left\{ 
    \begin{aligned} 
    &1+\mathcal{F}_{\text{\textit{abs}}}(P,P^*,\tau_{\text{\textit{abs}}}^1), \ \text{if} \ \mathcal{D}_{\text{\textit{norm}}} \leq \tau_{\text{\textit{norm}}} \\
    & \mathcal{F}_{\text{\textit{abs}}}(P,P^*,\tau_{\text{\textit{abs}}}^2), \text{else}
    \end{aligned}\right.\\
\end{aligned}
\end{equation}

where $w$ and $h$ denote screen resolution, while $\tau_{\text{\textit{norm}}}$, $\tau_{\text{\textit{abs}}}^1$ and $\tau_{\text{\textit{abs}}}^2$ represent thresholds. This approach offers two key benefits: \textbf{(\textit{i})} \textbf{it motivates the GUI agent to optimize along the path of minimal Euclidean distance when all predictions within a group align with the label}; \textbf{(\textit{ii})} \textbf{it rewards deviations from the nearest predictions when mismatches occur}. This multi-scale reward mechanism encourages the GUI agent to explore fine-grained screen details.

Via analysis of bad cases and guidelines $\mathbb{G}$, we notice that many incorrect predictions result from the GUI agent's lack of awareness of previous interaction histories, indicating weak short-term memory. Thus, we propose a model-based \textbf{Memory-Augmented Reward (MAR)} function. Our goal is to determine whether the explicit CoT $\mathcal{C}_{t,i}^*$ includes the agent's logical analysis of previous interactions. Let $r_{i}^{memory}$ denotes the MAR, which can be calculated as,
\begin{equation}
    r_{i}^{memory} = \mathcal{F}_{\text{MAR}}(\mathbb{P}_{\text{\textit{Memory}}}(\mathcal{C}_{t,i}^*,\mathscr{T}_{<t}^i))
\end{equation}

where $\mathbb{P}_{\text{\textit{Memory}}}$ represents the instruction for short-term memory verification and $\mathcal{F}_{\text{MAR}}(\cdot)$ denotes a model-based reward function\footnote{Considering the semantic diversity of the content and the weak flexibility of rule-based methods, we use Qwen3-235B-A22B~\cite{qwen3} as the memory reward judgment function.}. If CoT $\mathcal{C}_{t,i}^*$ contains historical information $\mathscr{T}_{<t}^i$, then $r_{i}^{memory}$=1, otherwise it is 0. Note that not all guidelines $\mathbb{G}$ contain clues to the context of historical interactions. Thus, compared with statically constraining the GUI agent to focus on historical interactions within the episode, our "guidelines + MAR" strategy can dynamically guide the agent to perform error-aware cognitive correction and enhance its short-term memory. Finally, we define the hybrid reward $r_i$ as follows, where $\gamma$ is a hyperparameter,
\begin{equation}
    \label{eq:final_reward}
    r_i = r_i^{\text{\textit{format}}}\times(r_i^{\text{\textit{action}}} + \gamma \times r_i^{\text{\textit{memory}}}) .
\end{equation}

\begin{figure}
    \centering
    \includegraphics[width=0.95\linewidth]{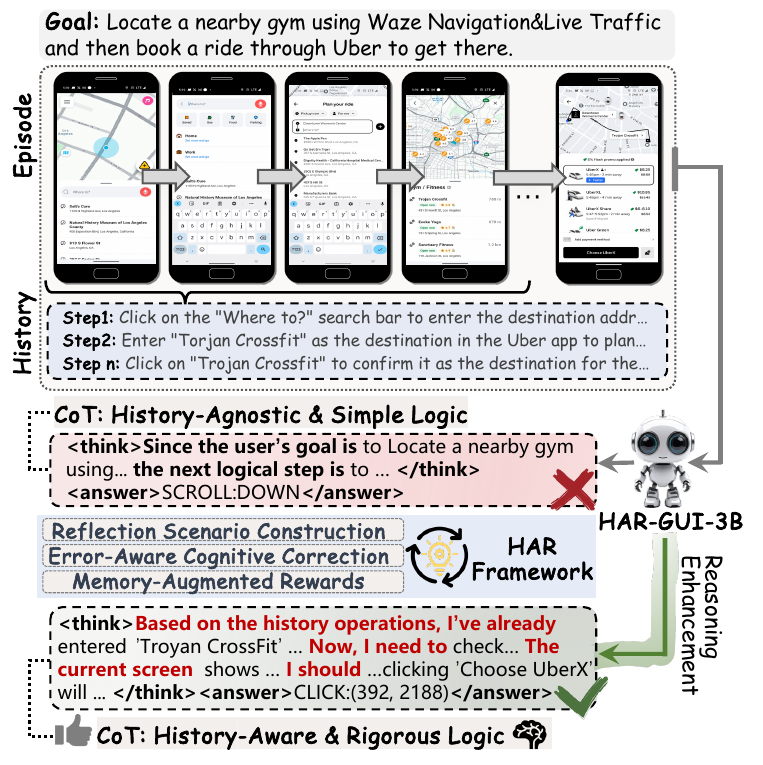}
    \caption{Short-term memory emergence and reasoning enhancement through HAR framework.}
    \label{fig:cot_details}
\end{figure}

After the first round of RL (Round-1 RL) in our constructed reflection scenario with tailored reward functions, we enhanced the short-term memory of the GUI agent in episodic reasoning, altering it from a history-agnostic and simplistic reasoning mode to a history-aware and rigorous reasoning mode. The effect of this stage is shown in Fig.\ref{fig:cot_details}.

\noindent\textbf{Round-2 RL.} Since the external episodic reasoning knowledge, i.e., tailored guidelines $\mathbb{{G}}$, is unavailable during the execution phase, we introduce Round-2 RL to align the execution format. Through Round-1 RL in the reflection scenario, the GUI agent enhances its capability to correct erroneous cognition, including improvements in short-term memory, action semantic understanding, and screen details perception. At this stage, instructions are converted to the inference-format to raise the difficulty of learning\footnote{Keeping the same instruction as when obtaining $\mathbb{D}_{his}$, and the output $\mathcal{O}_{t,i}^*$ only includes $\mathcal{A}_{t,i}^*$ and $\mathcal{C}_{t,i}^*$.}. Further, \textbf{we noticed that reflective training with episodic reasoning data alone weakens the GUI agent’s grounding capability} (\textit{as discussed in the ablation experiment, }Fig.\ref{fig:ablation}). To address this issue, we propose a \textbf{task mixing training strategy (TMTS)}, which is a multi-task RL approach that mixes grounding and GUI episodic reasoning tasks. For episodic reasoning task, we use the Round-1 RL hybrid reward function in Eq.\ref{eq:final_reward}; for grounding task, we use Eq.\ref{eq:click_reward}. See the appendix for details of each instruction template.

After completing Round-2 RL, we upgrade $\mathcal{M}_{\text{\textit{warm-up}}}$ to a reasoner $\mathcal{M}_{\text{\textit{HAR-GUI}}}$ (\model) with comprehensive GUI knowledge, which can serve as a GUI-tailored native model. Experiments demonstrate the advancement of $\mathcal{M}_{\text{\textit{HAR-GUI}}}$ across a range of GUI-related benchmarks.

\section{Experiments}\label{sec:experiments}
\subsection{Experimental Setup}
\noindent\textbf{Data Curation.} Overall, there are three training stages in the HAR framework. \textbf{(\textit{i})} We first distill GUI knowledge via the \textbf{warm-up SFT stage}, sampling 4k GUI understanding instances from MP-GUI, 20k grounding instances from OS-Atlas, and synthesizing 58k instances equipped with System-2 CoT alongside 100k Act2Sum entries from AITW, Mind2Web, and GUI-Odyssey. \textbf{(\textit{ii})} The \textbf{Round-1 RL stage} then conducts in the reflection scenario to perform error-aware cognitive correction. We synthesize 15k tailored guidance from hard samples to construct the reflection templates. \textbf{(\textit{iii})} Finally, the \textbf{Round-2 RL stage} employs our TMTS to enhance the GUI agent's screen detail perception and knowledge integration. We sample 15k grounding data from OS-Atlas and 15k instances from the pre-collected hard samples.

\noindent\textbf{Benchmarks.} We extensively evaluate the proposed \model\ on three types of widely used GUI-related benchmarks, \textbf{(\textit{i})} \textbf{GUI episodic reasoning benchmarks} include AITW~\cite{aitw}, Mind2Web~\cite{mind2web}, GUI-Odyssey~\cite{odyssey}, and a manually collected in-house data for OOD evaluation, \textbf{(\textit{ii})} \textbf{GUI grounding benchmarks} include ScreenSpot~\cite{cheng2024seeclick} and ScreenSpot-V2~\cite{li2025screenspot}, and \textbf{(\textit{iii})} \textbf{GUI understanding benchmarks} include ScreenQA (QA)~\cite{screenQA}, ScreenQA Short (QAS)~\cite{screenai}, Complex ScreenQA (CQA)~\cite{screenai}, WebSRC (WS)~\cite{chen2021websrc}, Screen2Words (S2W)~\cite{screen2words}, Taperception (TP)~\cite{tap} and Widget Captioning (WC)~\cite{widgetcaption}. 

\noindent\textbf{Implementation Details.}
Our \model\ is built upon Qwen2.5-VL-3B-Instruct and is trained using the proposed HAR framework.
We train the warm-up SFT stage for 1 epoch with a learning rate of 5e-6 and a global batch size of 256. For both Round-1 RL and Round-2 RL stages, we train for 2 epochs with a learning rate of 2e-6 and a batch size of 32. We set hyperparameters as $\tau_{\text{\textit{norm}}}$=0.1, $\tau_{abs}^1$=40, $\tau_{abs}^2$=200, and $\gamma$=0.2. LoRA with rank 64 and alpha 128 is applied to the LLM and Vision Backbone, using AdamW as optimizer. After training via the HAR framework, we perform post-training (SFT) of \model\ on GUI understanding~\cite{wang2025mp} and GUI episodic reasoning benchmarks individually with a learning rate of 2e-5 and batch size of 256, train each task for 4 epochs, and record the best results. Baselines use the same training settings as \model. Qwen2.5-VL-72B-Instruct serves as the teacher model for all data synthesis. All experiments are performed on 8 NVIDIA A100 80GB GPUs. 
\subsection{Main Results}
\noindent \textbf{Overall Performance.} We select AITW, Mind2Web and GUI-Odyssey to evaluate the agentic performance of \model. As shown in Tab.\ref{table:aitw-bench}, Tab.\ref{table:Mind2Web-benchmark}, and Tab.\ref{tab:odyssey}, \model\ consistently surpasses the current advanced methods, even those with far more parameters, \textit{e.g.} MP-GUI (8B), SeeClick (9.8B), and Qwen2.5-VL-7B. Under identical training settings, it delivers substantial gains over the foundational Qwen2.5-VL-3B. Further, when compared with GUI agents that share the same MLLM architecture, \textit{e.g.}, UI-R1-3B, GUI-R1-3B and InfiGUI-R1-3B, \model\ still leads by a clear margin.
\textbf{These results indicate that our HAR framework effectively enables domain-specific reasoning skills from hard samples in the constructed reflection scenario}, enabling reliable GUI automation.

\begin{table}\scriptsize
\renewcommand{\arraystretch}{.7} 
\setlength{\tabcolsep}{0.2mm}
\begin{tabular}{lcccccc}
\toprule
\textbf{Method}  & {\textbf{General}} & {\textbf{Install}} & {\textbf{G.Apps}} & {\textbf{Single}} & {\textbf{WebShop.}} & {\textbf{Overall}}      \\
\midrule
GPT-4V~\cite{yan2023gpt} & 41.7 & 42.6 & 49.8 & 72.8 & 45.7 & \cellcolor{mycolor}50.5 \\
OmniParser~\cite{lu2024omniparser} & 48.3 & 57.8 &51.6 & \underline{77.4} & 52.9 & \cellcolor{mycolor}57.7 \\
SeeClick~\cite{cheng2024seeclick} & 54.0 & 66.4 & 54.9 & 63.5 & 57.6 & \cellcolor{mycolor}59.3 \\
UI-R1-3B~\cite{lu2025ui} &  54.3	&63.6	&58.6	&68.2	&54.9	&\cellcolor{mycolor}59.9   \\
InternVL2-8B~\cite{chen2024far} & 58.1 & 65.3 & 56.8 & 68.7 & 61.1 &\cellcolor{mycolor}62.0\\
R-VLM~\cite{r-vlm} &59.9 &70.6& 59.6 &72.5 &61.7 &\cellcolor{mycolor}64.9\\
Qwen2.5-VL-3B~\cite{bai2025qwen2.5} & 61.2 & 69.8 & 62.9 & 70.8 & 62.4 & \cellcolor{mycolor}65.4\\
GUI-R1-3B~\cite{luo2025gui}& 59.2 & 68.9 & 64.5  & 71.9 & 63.2 & \cellcolor{mycolor}65.6 \\
InfiGUI-R1-3B~\cite{infigui-r1}& 62.6 & 72.3 & \underline{66.4} & 72.3 & 64.9 & \cellcolor{mycolor}67.7 \\
ShowUI~\cite{lin2025showui} & 63.5 & 72.3 & 66.0 & 72.3 & 65.8 & \cellcolor{mycolor}68.3 \\
MP-GUI~\cite{wang2025mp} & \underline{63.7} & \textbf{74.3} & 65.3 & 75.4 & \textbf{67.2} & \cellcolor{mycolor}\underline{69.2} \\
\midrule
\model\ & \textbf{63.8} & \underline{73.6} & \textbf{69.5} & \textbf{77.9} & \underline{66.1} & \cellcolor{mycolor}\textbf{70.2} \\
\bottomrule
\end{tabular}
\caption{Performance comparison on AITW. The evaluation metric used is step success rate (SSR).}
\label{table:aitw-bench}
\end{table}

\begin{table}\scriptsize
\renewcommand{\arraystretch}{.75} 
    \centering
    \setlength{\tabcolsep}{0.6mm}
    \begin{tabular}{@{}lccccccccc@{}}
        \toprule
        \multirow{2}{*}{\textbf{Method}}  &  \multicolumn{3}{c}{\textbf{Cross-Task}} & \multicolumn{3}{c}{\textbf{Cross-Website}} & \multicolumn{3}{c}{{\textbf{Cross-Domain}}} \\
        \cline{2-4} \cline{5-7} \cline{8-10}
        &  \textbf{Acc.}&\textbf{F1}&\textbf{SSR}&\textbf{Acc.}&\textbf{F1}&\textbf{SSR}&\textbf{Acc.}&\textbf{F1}&\textbf{SSR}\\ 
        \midrule
        InternVL2-8B~\cite{chen2024far} & 18.8	& 87.4	&16.7& 17.6& 85.8& 14.5 &13.9&	87.0	&12.0  \\
        CogAgent~\cite{cogagent} & 22.4 & 53.0 & 17.6 & 18.4 & 42.4 & 13.4 & 20.6 & 42.0 & 15.5 \\
        SeeClick~\cite{cheng2024seeclick} & 28.3 & 87.0 & 25.5 & 21.4 & 80.6 & 16.4 & 23.2 & 84.8 & 20.8 \\
        R-VLM~\cite{r-vlm} & 31.6& 88.0 &28.7 & 29.5 &84.9 &26.1& 26.7 &85.3 & 24.3\\
        OmniParser~\cite{lu2024omniparser} & \underline{42.4} & 87.6 & \underline{39.4} & 41.0 & 84.8 & 36.5 & \underline{45.5} & 85.7 & \underline{42.0} \\
        ShowUI~\cite{lin2025showui} & 39.7 & 88.0 & 36.9 & 41.0 & 83.6 & 34.2 & 38.9 & 85.3 & 34.1 \\
        UI-R1-3B~\cite{lu2025ui} &\underline{42.4}&	85.8	&36.8	&44.4&	83.1	&36.7&	43.0&	83.7&	36.3  \\
        Qwen2.5-VL-3B~\cite{bai2025qwen2.5} & 42.0 & 87.9 & 39.0 & 45.5 & 84.6 & 37.6 & 43.2 & 84.9 & 37.9 \\
        InfiGUI-R1-3B~\cite{infigui-r1}&  41.9&  86.5 &37.2  &  42.7& 84.9& 37.7 & 43.6& 83.1 & 38.2  \\
        GUI-R1-3B~\cite{luo2025gui}& 42.3 & 85.5 & 38.8  & \underline{45.8} & 84.2 & \underline{38.5} & 44.7 & 85.9& 38.9 \\
        MP-GUI~\cite{wang2025mp} & 42.1 & \underline{89.0} & 38.1 & 39.4 & \underline{87.1} & 32.9 & 37.6 & \underline{87.4} & 33.7 \\
        \midrule
        \model\ & \textbf{47.9}	&\textbf{89.6}	&\textbf{42.2}	&\textbf{49.1}	&\textbf{87.3}&	\textbf{41.2}	&\textbf{47.3}	&\textbf{88.3}	&\textbf{44.0}\\
        \bottomrule
    \end{tabular}
    \caption{Performance comparison on Mind2Web. We report element accuracy (Acc.), operation F1 (F1), and SSR.}
    \label{table:Mind2Web-benchmark}
\end{table}

\begin{table}[]\scriptsize
\renewcommand{\arraystretch}{.75}
    \centering
    \setlength{\tabcolsep}{0.36mm}
    \begin{tabular}{@{}lccccccc@{}}
    \toprule
    \textbf{Method} & \textbf{Tool} &\textbf{Info.} &\textbf{Shop.}& \textbf{Media}& \textbf{Social}& \textbf{M.Apps} & \textbf{Overall} \\
    \midrule
    GPT-4V~\cite{yan2023gpt}	&23.49	&20.16	&19.15	&16.92	&13.83	&19.02    &\cellcolor{mycolor}18.76\\
    GPT-4o~\cite{hurst2024gpt-4o}	&20.81	&16.28	&31.91	&15.38	&21.28	&16.67    &\cellcolor{mycolor}20.39\\
    Qwen2.5-VL-3B~\cite{bai2025qwen2.5}&  53.86	&43.44	&43.01	&44.74&	45.05	&45.72 & \cellcolor{mycolor}46.14 \\
    UI-R1-3B~\cite{lu2025ui} &  55.03&	43.81&	43.69&	45.17&	47.09&	45.46& \cellcolor{mycolor}46.71\\
    GUI-R1-3B~\cite{luo2025gui}&  57.46&	44.87	&44.71	&46.67	&49.42	&46.97& \cellcolor{mycolor}48.35\\
    InfiGUI-R1-3B~\cite{infigui-r1}&  60.60	&46.62	&45.19&	47.44	&53.48	&50.38& \cellcolor{mycolor}50.62\\
    Qwen2.5-VL-7B~\cite{bai2025qwen2.5} & \underline{70.24}	&\underline{57.24}	&\underline{49.28}	&\underline{58.34}	&\underline{60.32}	&\underline{54.93} & \cellcolor{mycolor}\underline{58.39} \\
    \midrule
    \model\ &  \textbf{74.62}& \textbf{58.53}& \textbf{51.32} &\textbf{ 62.19}& \textbf{65.51}& \textbf{61.70}& \cellcolor{mycolor}\textbf{62.31}\\
    \bottomrule
    \end{tabular}
    \caption{Performance on GUI-Odyssey. The metric is SSR.}
    \label{tab:odyssey}
\end{table}

\begin{table}[t!]
\centering
    \setlength{\tabcolsep}{1mm}
    \resizebox{\linewidth}{!}{
    \begin{tabular}{@{}lccccccc@{}}
        \toprule
         \multirow{2}{*}{\textbf{Method}} & \multicolumn{2}{c}{\textbf{Mobile}} & \multicolumn{2}{c}{\textbf{Desktop}} & \multicolumn{2}{c}{\textbf{Web}} &  \multirow{2}{*}{Avg.} \\
        \cline{2-3} \cline{4-5} \cline{6-7}
        & Text & Icon & Text & Icon & Text & Icon & \\
        \midrule
        GPT-4o~\cite{hurst2024gpt-4o} & 30.5 & 23.2 & 20.6 & 19.4 & 11.1 & 7.8 & \cellcolor{mycolor}18.8 \\
        CogAgent~\cite{cogagent}&67.0 &24.0 &74.2 &20.0 &70.4 &28.6 &\cellcolor{mycolor}47.4\\
        SeeClick~\cite{cheng2024seeclick}&	78.4	&50.7	&70.1	&29.3	&55.2	&32.5&\cellcolor{mycolor}	55.1 \\
        R-VLM~\cite{r-vlm} & 85.0 &61.1 &81.4 &52.8 &66.5& 51.4 &\cellcolor{mycolor}66.3\\
        MP-GUI~\cite{wang2025mp} & 86.8 & 65.9 & 70.8 & 56.4 & 58.3 & 46.6 & \cellcolor{mycolor}64.1 \\
        UGround-7B~\cite{gou2024navigating} & 82.8 & 60.3 & 82.5 & 63.6 & 80.4 & 70.4 & \cellcolor{mycolor}73.3 \\
        ShowUI~\cite{lin2025showui} & 92.3 & \underline{75.5} & 76.3 & 61.1 & 81.7 & 63.6 & \cellcolor{mycolor}75.1 \\
        Qwen2.5-VL-7B~\cite{bai2025qwen2.5} & \underline{93.8}	&72.5 &	87.6 & 65.7 &	88.7&	70.4&\cellcolor{mycolor} 79.8\\
        UI-R1-3B~\cite{lu2025ui} & -- & -- & 90.2 & 59.3 & 85.2 & 73.3 & \cellcolor{mycolor}-- \\
        GUI-R1-3B~\cite{luo2025gui} & -- & -- & \underline{93.8} & 64.8 & \underline{89.6} & 72.1 & \cellcolor{mycolor}-- \\
        
        UI-TARS-2B~\cite{qin2025ui} & 93.0 & \underline{75.5} & \textbf{94.3} & \underline{68.6} & 84.3 & \textbf{74.8} & \cellcolor{mycolor}82.3 \\
        OS-Atlas-7B~\cite{wu2024atlas}  & 93.0 & 72.9 & 91.8 & 62.9 & \textbf{90.9} & \underline{74.3} & \cellcolor{mycolor}\underline{82.5} \\
        \midrule
        \model\ & \textbf{94.5} & \textbf{81.0} & \underline{93.8} & \textbf{70.8} & 85.6 & 73.8 & \cellcolor{mycolor}\textbf{83.3} \\
        \bottomrule
    \end{tabular}
    }
    \caption{Performance comparison on ScreenSpot.}
    \label{tab:screenspot}
\end{table}
\begin{table}[h!]\scriptsize
\renewcommand{\arraystretch}{.75}
\centering
    \setlength{\tabcolsep}{0.8mm}
    \begin{tabular}{lccccccc}
        \toprule
        \multirow{2}{*}{\textbf{Method}} & \multicolumn{2}{c}{\textbf{Mobile}} & \multicolumn{2}{c}{\textbf{Desktop}} & \multicolumn{2}{c}{\textbf{Web}} & \multirow{2}{*}{Avg.} \\
        \cline{2-3} \cline{4-5} \cline{6-7}
        & Text & Icon & Text & Icon & Text & Icon & \\ 
        \midrule
        SeeClick~\cite{cheng2024seeclick} & 78.4 & 50.7 & 70.1 & 29.3 & 55.2 & 32.5 & \cellcolor{mycolor}55.1 \\
        GPT-4o + SeeClick~\cite{cheng2024seeclick} & 85.2 & 58.8 & 79.9 & 37.1 & 72.7 & 30.1 & \cellcolor{mycolor}63.6 \\
        OS-Atlas-4B~\cite{wu2024atlas} & 87.2 & 59.7 & 72.7 & 46.4 & 85.9 & 63.1 & \cellcolor{mycolor}71.9 \\
        GPT-4o + OS-Atlas-4B~\cite{wu2024atlas} & 95.5 & 75.8 & 79.4 & 49.3 & 90.2 & 66.5&  \cellcolor{mycolor}79.1 \\
        InternVL3-8B~\cite{zhu2025internvl3}& -- &-- &-- & --& --&-- & \cellcolor{mycolor} 81.4\\ 
        Qwen2.5-VL-3B~\cite{bai2025qwen2.5} & 95.0	&80.1	&90.2&	64.3&	88.0&	70.4	&\cellcolor{mycolor}81.3\\
        OS-Atlas-7B~\cite{wu2024atlas} & 95.2 & 75.8 & 90.7 & 63.6 & \underline{90.6} & \underline{77.3} & \cellcolor{mycolor}84.1 \\
        UI-TARS-2B~\cite{qin2025ui} & 95.2 & 79.1 & 90.7 & 68.6 & \underline{90.6} & \underline{77.3} & \cellcolor{mycolor}84.7 \\
        UI-R1-3B~\cite{lu2025ui} &\underline{96.2}	&\textbf{84.3}	&\underline{92.3}	&63.6	&89.2 &	75.4 & \cellcolor{mycolor}\underline{85.4}\\
        \midrule
        \model\ & \textbf{96.5} & \underline{81.0} & \textbf{95.4} & \textbf{76.5} & 88.8 & \textbf{78.8} & \cellcolor{mycolor}\textbf{86.2}\\
        \bottomrule
    \end{tabular}
    \caption{Performance comparison on ScreenSpot-V2.}
    \label{tab:screenspot_v2}
\end{table}
\noindent\textbf{Grounding.} Grounding capability is crucial for GUI automation~\cite{cheng2024seeclick}, as it determines whether the GUI agent can accurately execute click-based actions on the screen. We compare the grounding performance of \model\ and the current advanced methods on ScreenSpot and ScreenSpot-V2. As shown in Tab.\ref{tab:screenspot} and Tab.\ref{tab:screenspot_v2}, \model\ delivers the leading results. We attribute this reliable screen grounding performance to the TMTS (Sect.\ref{sec:2round_RL}) used in Round-2 RL training stage\footnote{We find integrating the grounding task into episodic reasoning improves both tasks compared with training them sequentially.}. \textbf{The mixture of the grounding task further enhances the GUI agent in screen detail perception and GUI knowledge acquisition}.

\begin{figure*}[]
\begin{minipage}[]{1.2\columnwidth}
\renewcommand{\arraystretch}{0.9}
    \begin{minipage}{\linewidth}
    \centering
    \setlength{\tabcolsep}{1mm}
    \small
\begin{tabular}{ccc|ccc}
\toprule
\multicolumn{3}{c}{\textbf{Takeout}} & \multicolumn{3}{|c}{\textbf{ Repast}} \\
\#Tasks & \#Steps & Avg. Steps & \#Tasks & \#Steps & Avg. Steps \\ \hline
103       & 1,031    & 10.11     & 108       & 1,015    & 9.49      \\ \hline

\multicolumn{3}{c}{\textbf{Finance}} & \multicolumn{3}{|c}{\textbf{Insurance}} \\
\#Tasks & \#Steps & Avg. Steps & \#Tasks & \#Steps & Avg. Steps \\ \hline
102       & 489     & 4.82      & 102       & 503     & 4.98      \\ \bottomrule
\end{tabular}

\captionof{table}{In-house GUI episode data statistics cover 4 categories of widely used Chinese mobile app scenarios, all with manual annotations.}
\label{tab:ood_stastic}
    \end{minipage}
    \hfill
    \begin{minipage}{\linewidth}
\centering
\scriptsize
\renewcommand{\arraystretch}{0.8} 
\setlength{\tabcolsep}{1mm}
\begin{tabular}{@{}l *{8}{c} @{}}
\toprule
 \multirow{2}{*}{\textbf{Method}} & \multicolumn{2}{c}{\textbf{Takeout}} & \multicolumn{2}{c}{\textbf{Repast}} & \multicolumn{2}{c}{\textbf{Finance}} & \multicolumn{2}{c}{\textbf{Insurance}} \\
\cmidrule(lr){2-3} \cmidrule(lr){4-5} \cmidrule(lr){6-7} \cmidrule(lr){8-9} 
& \textbf{SSR} & \textbf{SR} & \textbf{SSR} & \textbf{SR} & \textbf{SSR} & \textbf{SR} & \textbf{SSR} & \textbf{SR} \\
\midrule
Qwen2.5-VL-72B~\cite{bai2025qwen2.5} & \cellcolor{mygreen}\underline{80.60} & \cellcolor{mygreen}\underline{17.65} & \cellcolor{mypink}\textbf{85.02} & \cellcolor{mypink}\textbf{20.56} & \cellcolor{mypink}\textbf{86.91} & \cellcolor{mypink}\textbf{57.43} & \cellcolor{mypink}\textbf{69.98} & \cellcolor{mypink}\textbf{21.78} \\
Gemini-2.5-Pro~\cite{Gemini2025}& \cellcolor{mygray}71.96	&4.76	&68.83	&\cellcolor{mygray}6.74	&48.81	&5.62	&57.83	&10.23  \\
UI-TARS-7B-DPO~\cite{qin2025ui}& 37.65	&0.00	&39.47	&0.00	&70.80	&\cellcolor{mygreen}\underline{48.80}	&36.84	&11.54\\
MiMo-VL-7B-RL~\cite{mimo-vl}	&56.74	&0.98	&64.24	&3.74	&\cellcolor{mygray}73.41	&\cellcolor{mygray}35.64	&\cellcolor{mygray}64.21	&\cellcolor{mygray}14.85\\
Qwen2.5-VL-7B~\cite{bai2025qwen2.5}  & 46.27 & 0.00  & 69.26 & 3.74  & 59.71 & 20.79 & 56.66 & \cellcolor{mygreen}\underline{16.83} \\
GUI-R1-3B~\cite{luo2025gui}&71.21	&\cellcolor{mygray}10.7	&\cellcolor{mygray}70.49	&5.77	&69.99	&30.01	&63.60	&14.78 \\
UI-R1-3B~\cite{lu2025ui}&59.60&	1.00&	67.31	&4.76	&61.15 &	19.80	&58.63 &	9.90\\
Qwen2.5-VL-3B~\cite{bai2025qwen2.5}&	51.21	&0.00	&62.01&	0.00	&57.08	&18.19	&50.51	&8.24 \\
InfiGUI-R1-3B~\cite{infigui-r1}&	61.90&	0.00&	69.11&	3.99	&51.74	&8.77&	45.69&	7.55\\
\hline
\model\  &\cellcolor{mypink}\textbf{ 82.76 }& \cellcolor{mypink}\textbf{24.30} & \cellcolor{mygreen}\underline{77.69} & \cellcolor{mygreen}\underline{11.76} & \cellcolor{mygreen}\underline{76.50}  & 35.60  & \cellcolor{mygreen}\underline{69.19} & \cellcolor{mypink}\textbf{21.78} \\
\bottomrule
\end{tabular}
\captionof{table}{Zero-Shot comparison in OOD scenarios. We report SSR and episode-wise success rate (SR). \colorbox{mypink}{Optimal}, \colorbox{mygreen}{suboptimal} and \colorbox{mygray}{third-best}.}
\label{tab:ood_performance}
    \end{minipage}
\end{minipage}
\hfill
\begin{minipage}[]{0.88\columnwidth}
    \centering
    \includegraphics[width=1.0\linewidth]{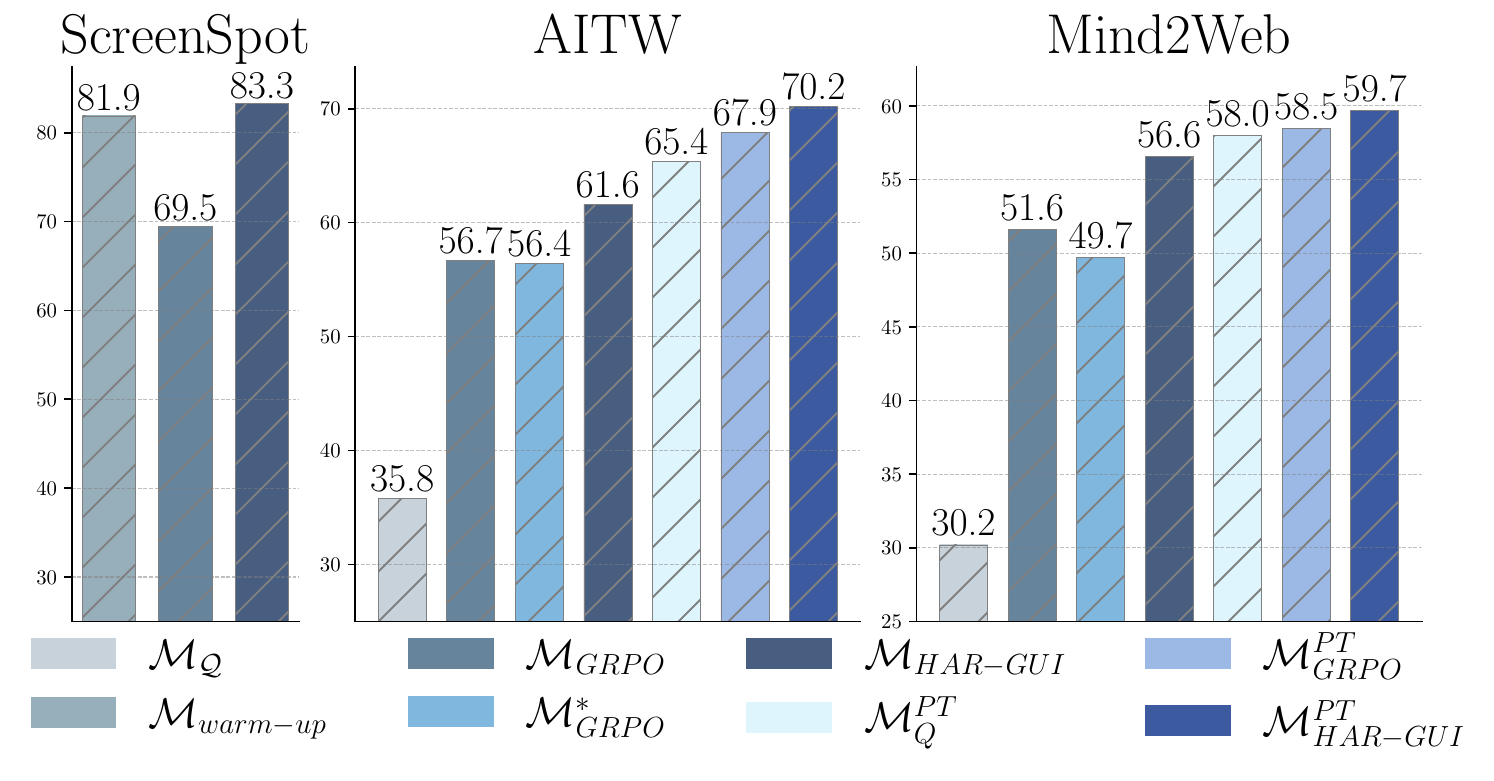}
    \caption{Effectiveness of HAR framework. $\mathcal{M}_{\mathcal{Q}}$ is Qwen2.5-VL-3B zero-shot results. $\mathcal{M}_{\mathcal{\text{\textit{GRPO}}}}$ refers to the method keeping the same settings as $\mathcal{M}_{\mathcal{\text{\textit{HAR-GUI}}}}$, but excluding the reflection scenario (using the inference-format instruction), MAR, and TMTS. $\mathcal{M}_{\mathcal{\text{\textit{Q}}}}^{PT}$, $\mathcal{M}_{\mathcal{\text{\textit{GRPO}}}}^{PT}$ and $\mathcal{M}_{\mathcal{\text{\textit{HAR-GUI}}}}^{PT}$ are the post-training results on each benchmark for $\mathcal{M}_{\mathcal{\text{\textit{Q}}}}$, $\mathcal{M}_{\mathcal{\text{\textit{GRPO}}}}$ and $\mathcal{M}_{\mathcal{\text{\textit{HAR-GUI}}}}$, respectively. Compared with $\mathcal{M}_{\mathcal{\text{\textit{GRPO}}}}$, $\mathcal{M}_{\mathcal{\text{\textit{GRPO}}}}^*$ mandates that the agent focuses on historical interaction context of the episode in the instructions.}
    \label{fig:ablation}
\end{minipage}
\end{figure*}
 
\noindent\textbf{OOD Evaluation.} To evaluate the OOD generalization of current SOTA methods, we develop a challenging Chinese mobile automation benchmark (Tab.\ref{tab:ood_stastic}). We manually collect and annotate 415 tasks from Alipay mini programs ({\tt CLICK}-only action space). As reported in Tab.\ref{tab:ood_performance}, \model\ significantly outperforms other advanced methods with comparable parameter sizes. \textbf{This remarkable performance mainly stems from its stable short-term memory and its reliable screen perception capability.}

\noindent\textbf{GUI Understanding.} In this part, we compare the basic GUI understanding effect of \model. We select the comprehensive benchmarks collected by MP-GUI, which include screen analysis (WC and TP), screen question-answering (QA, CQA, QAS, and WS) and screen summarization (S2W). In Tab.\ref{table:gui_under_bench}, \model\ outperforms advanced methods on most GUI-oriented tasks with fewer parameters. Compared to the foundational Qwen2.5-VL-3B, our \model\ shows an overall gain of 3.07 points with the same training settings. Compared with methods with larger parameters, \model\ is still competitive. In particular, our method surpasses the current SOTA GUI-specific method MP-GUI (8B) on S2W, WS, QA and QAS benchmarks, and UI-TARS-1.5-7B on S2W and TP benchmarks. \textbf{Such results demonstrate that \model\ can implicitly learn multi-grained screen knowledge through error-aware cognitive correction and tailored training recipe.}

\begin{table}[h]\scriptsize
\renewcommand{\arraystretch}{.75}
\centering
\setlength{\tabcolsep}{0.9mm}
\begin{tabular}{lccccccc}
\toprule
 \textbf{Method} & \textbf{WC} & \textbf{S2W} & \textbf{TP} & \textbf{WS} & \textbf{QA} & \textbf{QAS} & \textbf{CQA} \\
\midrule
Llama 3.2-V (11B)~\cite{Llama3}  & 113.6 & 108.8  & 83.4 & 87.0 & 88.4 & {\textbf{91.6}} & 74.6 \\
CogAgent (18B)~\cite{cogagent}  &136.2& 115.0& {\textbf{88.4}}& 63.1& 85.3& 74.6& 65.1  \\
UI-TARS-2B-SFT~\cite{qin2025ui}& 125.8& 115.0& 80.5 & 88.5& 86.2& 90.1& 80.3 \\
InternVL2 (8B)~\cite{chen2024far} & 140.6& 115.2& 86.7& 89.7 & 84.2& 89.2& {\underline{82.4}} \\
InfiGUI-R1-3B~\cite{infigui-r1} &142.4	&116.9&	85.2&	91.4&	88.1&	86.4	&77.6 \\
GUI-R1-3B~\cite{luo2025gui} & 141.7	&117.5	&84.9	&92.2	&88.3	&87.2	&79.0 \\
Qwen2.5-VL-3B~\cite{bai2025qwen2.5}  &133.3&	117.9	&86.5	&90.1	&87.4&	90.3&	79.6 \\
UI-R1-3B~\cite{lu2025ui} & 142.6	&{118.4}	&87.0	& {93.0}	&{\underline{89.1}}	&90.4	&78.8 \\
MP-GUI (8B)~\cite{wang2025mp} & {\textbf{151.0}} & {118.4} & {\underline{88.2}} & 89.2 & {88.6} & {90.5} & {\textbf{84.3}}\\
UI-TARS-1.5-7B~\cite{qin2025ui}& {\underline{147.4}}	&{\underline{118.7}} &87.1 & {\textbf{94.1}} & {\textbf{89.5}} & {\underline{91.2}} & {82.1}\\
\midrule
\model\ & {143.9}	&{\textbf{119.1}}	&{87.5} &	{\underline{93.5}}	&{\textbf{89.5}}	&91.0	&{82.1} \\
\bottomrule
\end{tabular}
\caption{Performance comparison on GUI understanding benchmarks. We employ CIDEr scores to assess WC and S2W, SQuAD F1 scores for QAS, CQA and WS, ROUGE-L scores for QA, and F1 values for TP.}
\label{table:gui_under_bench}
\end{table}

\subsection{Ablation Study}
We conduct an ablation study to validate our HAR framework.
Starting from $\mathcal{M}_{\text{\textit{warm-up}}}$, we tune GUI agents with the HAR framework to produce $\mathcal{M}_{\text{\textit{HAR-GUI}}}$ and with vanilla GRPO\footnote{Without (\textit{w}/\textit{o}) reflection scenario, MAR, and TMTS.} to yield $\mathcal{M}_{\textit{GRPO}}$ and $\mathcal{M}_{\textit{GRPO}}^*$. We then compare their screen grounding on ScreenSpot and episodic reasoning on AITW and Mind2Web.
Further, we keep the same settings to conduct post-training on AITW and Mind2Web using $\mathcal{M}_{\textit{Q}}$, $\mathcal{M}_{\textit{GRPO}}$ and $\mathcal{M}_{\textit{HAR-GUI}}$ as initial checkpoints to verify the generalizability of our method.

Fig.\ref{fig:ablation} illustrates that: \textbf{(\textit{i})} A comparison of $\mathcal{M}_{\textit{GRPO}}$ and $\mathcal{M}_{\textit{warm-up}}$ on ScreenSpot reveals that \textbf{training solely with episodic reasoning weakens the agent’s grounding capability}, whereas our TMTS mitigates this degradation. \textbf{(\textit{ii})} Comparing $\mathcal{M}_{\textit{GRPO}}^*$ and $\mathcal{M}_{\textit{GRPO}}$, it is evident that a prompt-constrained focus on historical context can lead to performance degradation. We argue that \textbf{the agent's reasoning mode should not be constrained, but autonomously shaped and adapted during self-evolution}. \textbf{(\textit{iii})} On AITW and Mind2Web, $\mathcal{M}_{\textit{HAR-GUI}}$ outperforms $\mathcal{M}_{\textit{GRPO}}$, and in identical post-training settings, $\mathcal{M}_{\textit{HAR-GUI}}^{PT}$ surpasses both $\mathcal{M}_{\textit{GRPO}}^{PT}$ and $\mathcal{M}_{\textit{Q}}^{PT}$. These results demonstrate the effectiveness of our HAR framework.
Further, we observed that both $\mathcal{M}_{\textit{GRPO}}$ and $\mathcal{M}_{\textit{GRPO}}^*$ still exhibit weak short-term memory after RL training using the inference-format instruction. In contrast, the emerging history-aware and rigorous reasoning mode of $\mathcal{M}_{\textit{HAR-GUI}}$ is driven by the HAR framework, primarily via \textbf{tailored guidance synthesis}, \textbf{cognitive correction in the reflection scenario}, and \textbf{MAR}.

\section{Conclusion} \label{sec:conclusion}
We propose the HAR framework to enhance the reasoning of GUI agents via reflective training, especially equipping the agent with stable short-term memory for episodic reasoning. The framework consists mainly of constructing a reflective learning scenario, synthesizing tailored correction guidelines, and designing a hybrid RL reward function. Via HAR, we develop a native model \model, which integrates reliable performance to handle GUI-oriented tasks.

\section*{Acknowledgments}
This work was supported by the National Natural Science Foundation of China (Grant No.62372408).This work was supported by Ant Group Research Fund.

\bibliography{aaai2026}
\clearpage
\appendix
\section{Appendix}

\subsection{Instruction Template Details}\label{sec:instruction_details}
In this section, we detail the instruction templates used in each stage of the work introduced earlier, including data synthesis, reflection scenario construction, inference-format instruction for the GUI episode reasoning task, and downstream tasks. For the GUI understanding benchmark, the instruction template we use is consistent with MP-GUI. For methods with output coordinates ranging from 0 to 1000 (\textit{e.g.}, the UI-TARS), we scaled both the instruction template and output coordinates accordingly to ensure alignment with the original approach. For the listed baseline methods, we refer to the official documentation and make the requisite format adjustments to achieve action alignment. Detailed instructions are shown in the colored box below. Our experiments noticed that \model\ reliably follows instructions, enabling researchers to refer to these predefined instructions for various customized zero-shot inference and post-tuning GUI-oriented tasks.
\begin{table*}[ht]
    \centering
    \arrayrulecolor{black}
    \begin{tabular}{|p{5cm}|p{11cm}|}
        \hline
        \rowcolor{gray!10}
        \textbf{Action} & \textbf{Description} \\ \hline
        CLICK:(x, y) & Click on the element at the coordinate point (x,y) on the screen, e.g.,  CLICK:(1980,224). \\ \hline
        TYPE: typed text & An action of typing a piece of text, e.g., TYPE:"Macbook-Pro 16G Black". \\ \hline
        COMPLETE & The goal has been completed in the current screen state. \\ \hline
        SCROLL:UP/DOWN/LEFT/RIGHT & Scroll in a specific direction, e.g., SCROLL:UP \\ \hline
        BACK & Go back to the previous screen, e.g, BACK.\\ \hline
        HOME & Go to the home screen, e.g., HOME. \\ \hline
        ENTER & Press the ENTER key to submit input content. \\ \hline
        TYPE:(x, y, typed text) & An action of typing a piece of text at the coordinate point (x,y), e.g., TYPE:(208,1082,Macbook-Pro 16G Black). \\ \hline
        SELECT:(x, y, option) & Opens the Select Menu or Drop-Down List at coordinate (x, y) and selects the option you specify, e.g., SELECT:(59,892,Chicago). \\ \hline
        \textnormal{LONG\_PRESS}:(x, y) & Long press at a specific point (x,y) on the screen, e.g., \textnormal{LONG\_PRESS}:(345,2218). \\ \hline
        \textnormal{PRESS\_RECENT} & The action to go to the previous app.\\ \hline
        IMPOSSIBLE & No action can be taken to achieve the goal based on current screen. \\ \hline
    \end{tabular}
    \caption{Unified action space of \model. }
    \label{tab:action_space}
\end{table*}
\subsection{Action Space}\label{sec:action_space}
Tab.\ref{tab:action_space} lists our predefined action space for the GUI episode reasoning task, which contains the most commonly used interactive operations of smart devices. The actions used in the AITW benchmark include "CLICK: (x, y)", "TYPE: typed text", "COMPLETE", "SCROLL: UP/DOWN/LEFT/RIGHT", "BACK", "HOME", and "ENTER". The actions used in the Mind2Web benchmark include "CLICK: (x, y)", "TYPE: (x, y, typed text)" and "SELECT: (x, y, option)". The actions used in the GUI-Odyssey benchmark include "CLICK: (x, y)", "TYPE: typed text", "COMPLETE", "SCROLL: UP/DOWN/LEFT/RIGHT", "BACK", "HOME", "\textnormal{LONG\_PRESS}:(x, y)", "\textnormal{PRESS\_RECENT}" and "IMPOSSIBLE". For AITW and Mind2Web, we use the same data processing as SeeClick.

\subsection{Case Study}\label{sec:case_study}
\subsubsection{Details of usage data}
\begin{figure*}
    \centering
    \includegraphics[width=1.0\linewidth]{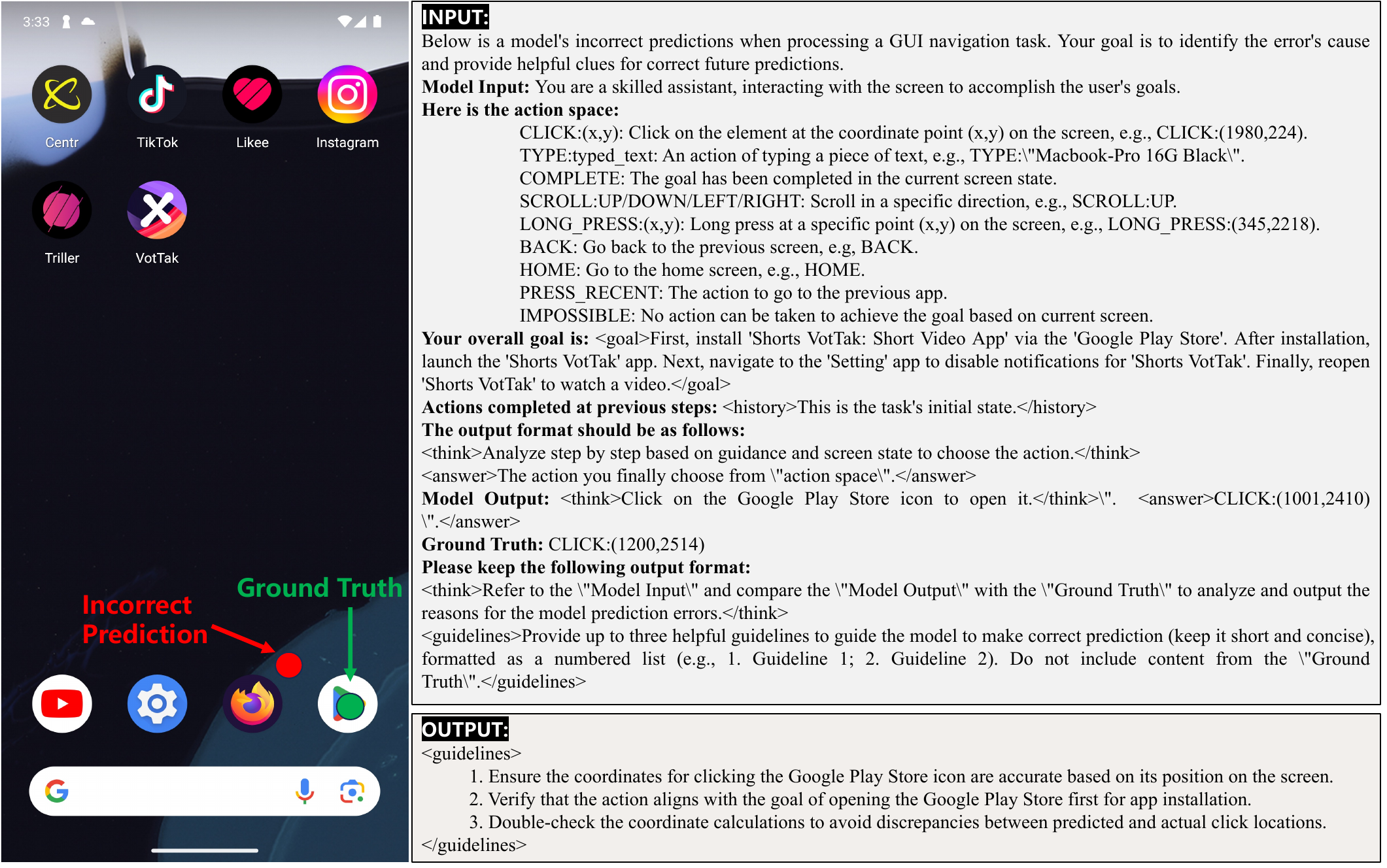}
    \caption{Case of guidance synthesis for error correction. The results are generated by the teacher model Qwen2.5-VL-72B-Instruct.}
    \label{fig:guideline-demo1}
\end{figure*}

\begin{figure*}
    \centering
    \includegraphics[width=0.65\linewidth]{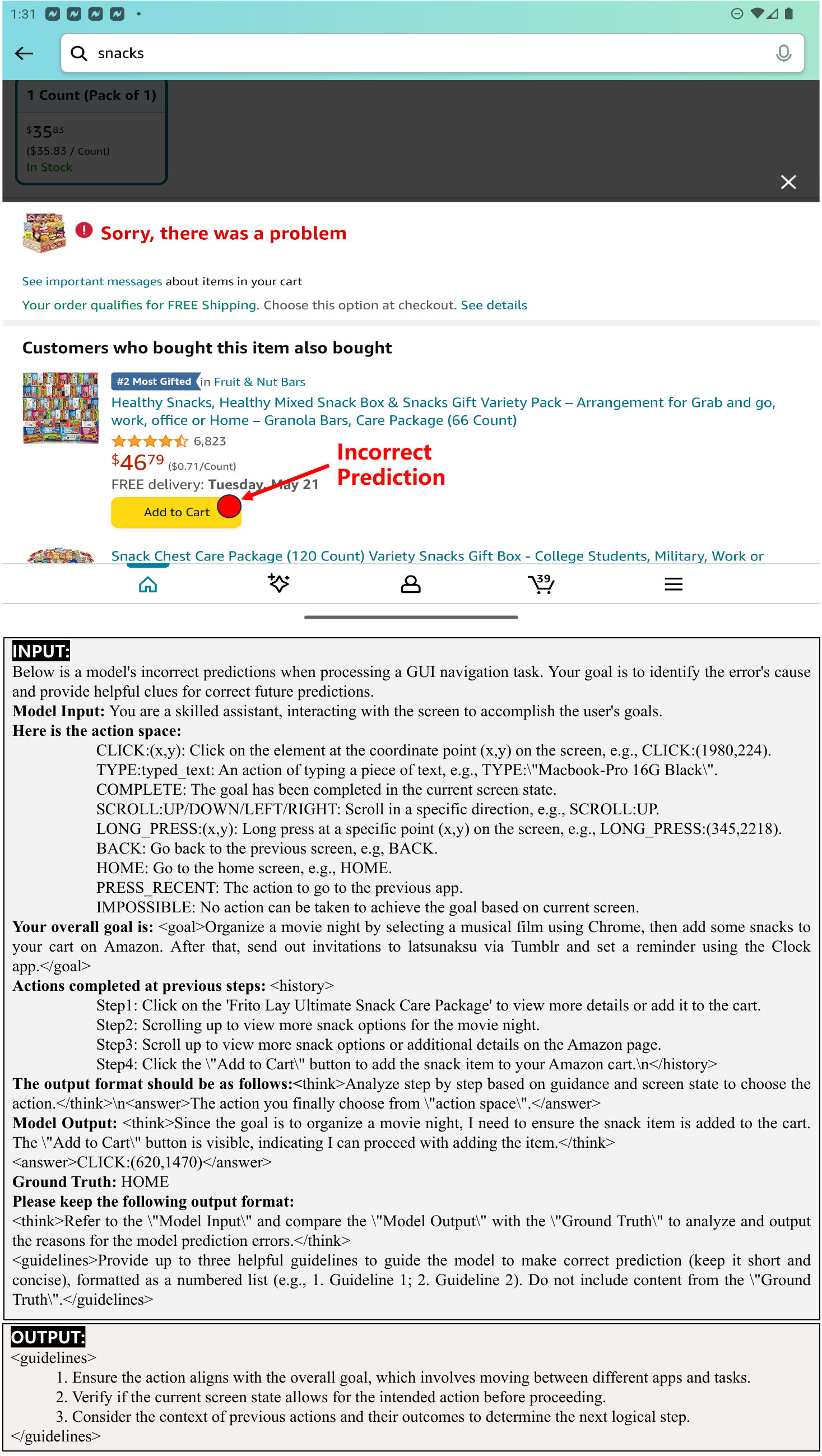}
    \caption{Case of guidance synthesis for error correction. The results are generated by the teacher model Qwen2.5-VL-72B-Instruct.}
    \label{fig:guideline-demo2}
\end{figure*}

\begin{figure*}
    \centering
    \includegraphics[width=1.0\linewidth]{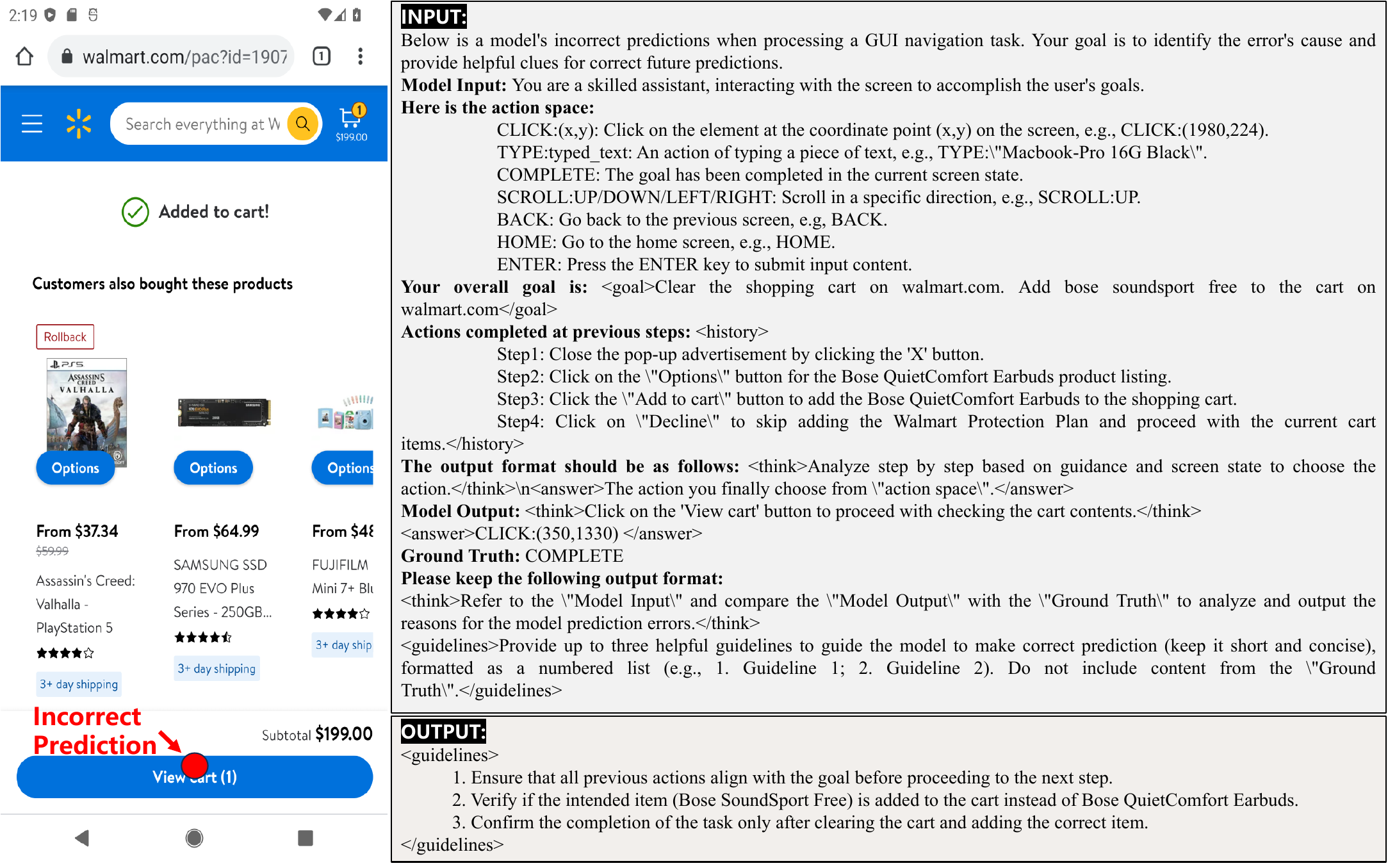}
    \caption{Case of guidance synthesis for error correction. The results are generated by the teacher model Qwen2.5-VL-72B-Instruct.}
    \label{fig:guideline-demo3}
\end{figure*}

\begin{figure*}
    \centering
    \includegraphics[width=1.0\linewidth]{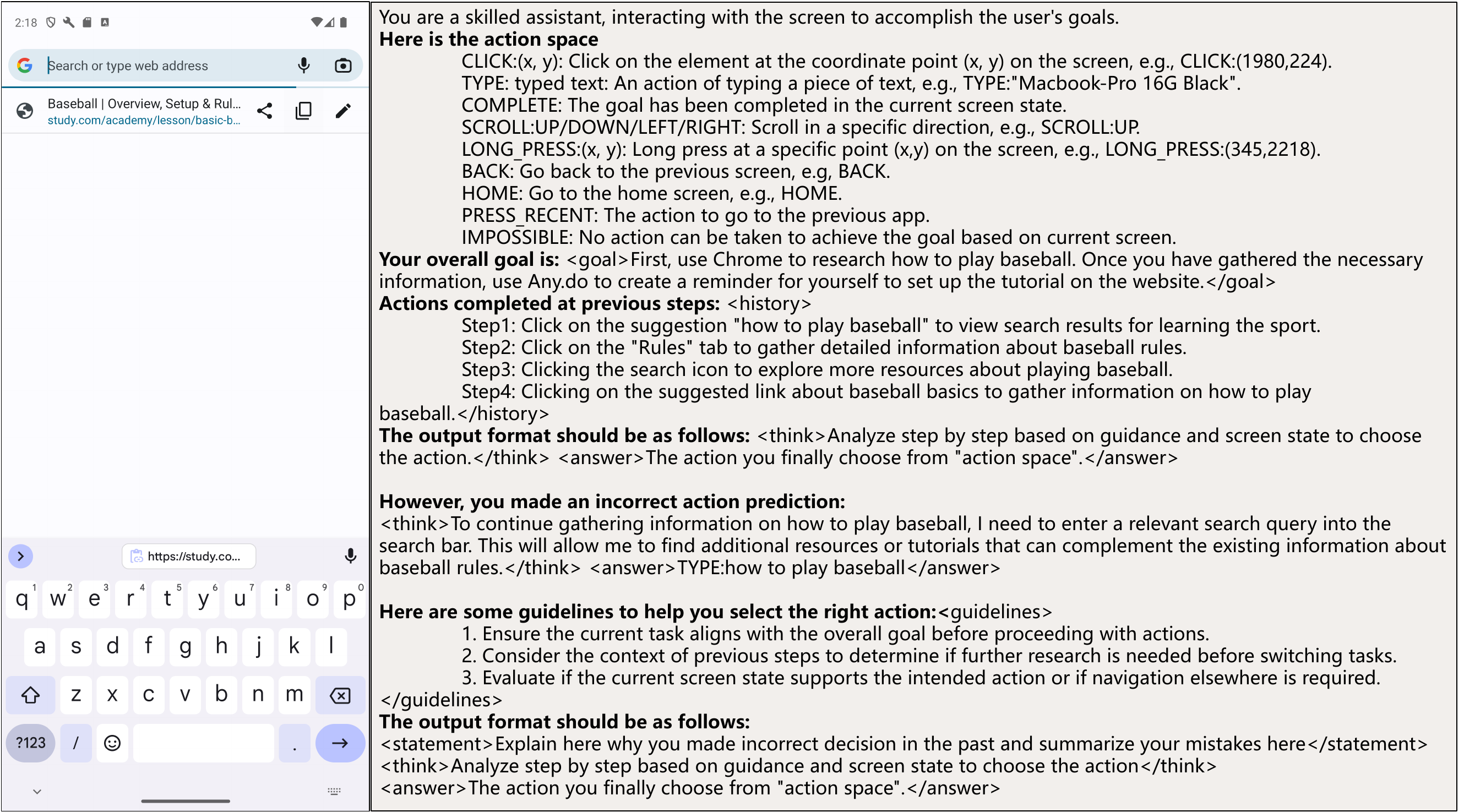}
    \caption{An example of data used for error-aware cognitive correction in the reflection scenario.}
    \label{fig:reflection-sample}
\end{figure*}

In this part, we present the details of the data used in each stage of the HAR framework. We employ a teacher model (Qwen2.5-VL-72B-Instruct) to conduct error analysis on hard samples and synthesize correction guidelines. These guidelines, serving as external episodic reasoning knowledge, assist the GUI agent rectify its misinterpretations within the constructed reflection scenario. Fig.\ref{fig:guideline-demo1}, Fig.\ref{fig:guideline-demo2} and Fig.\ref{fig:guideline-demo3} provide examples of these correction guidelines. When constructing the reflection-format instruction template, we map each entry based on its corresponding screenshot image ID. Fig.\ref{fig:reflection-sample} details the sample used in the Round-1 RL training stage.

\subsubsection{OOD scenario performance analysis}
In this part, we analyze a complete episodic reasoning result in our manually annotated Chinese OOD benchmark.
Considering that each baseline method has different action space during development, yet all define the {\tt CLICK} action. Thus, we only selected the {\tt CLICK}-only episodes in our in-house data for a fair evaluation, that is, there is only one {\tt CLICK} action in the action space (Tab.\ref{tab:action_space}), which is also the most commonly used action by most users when using mobile phones. All data is manually collected and labeled within the Alipay app.

\begin{figure*}
    \centering
    \includegraphics[width=1.0\linewidth]{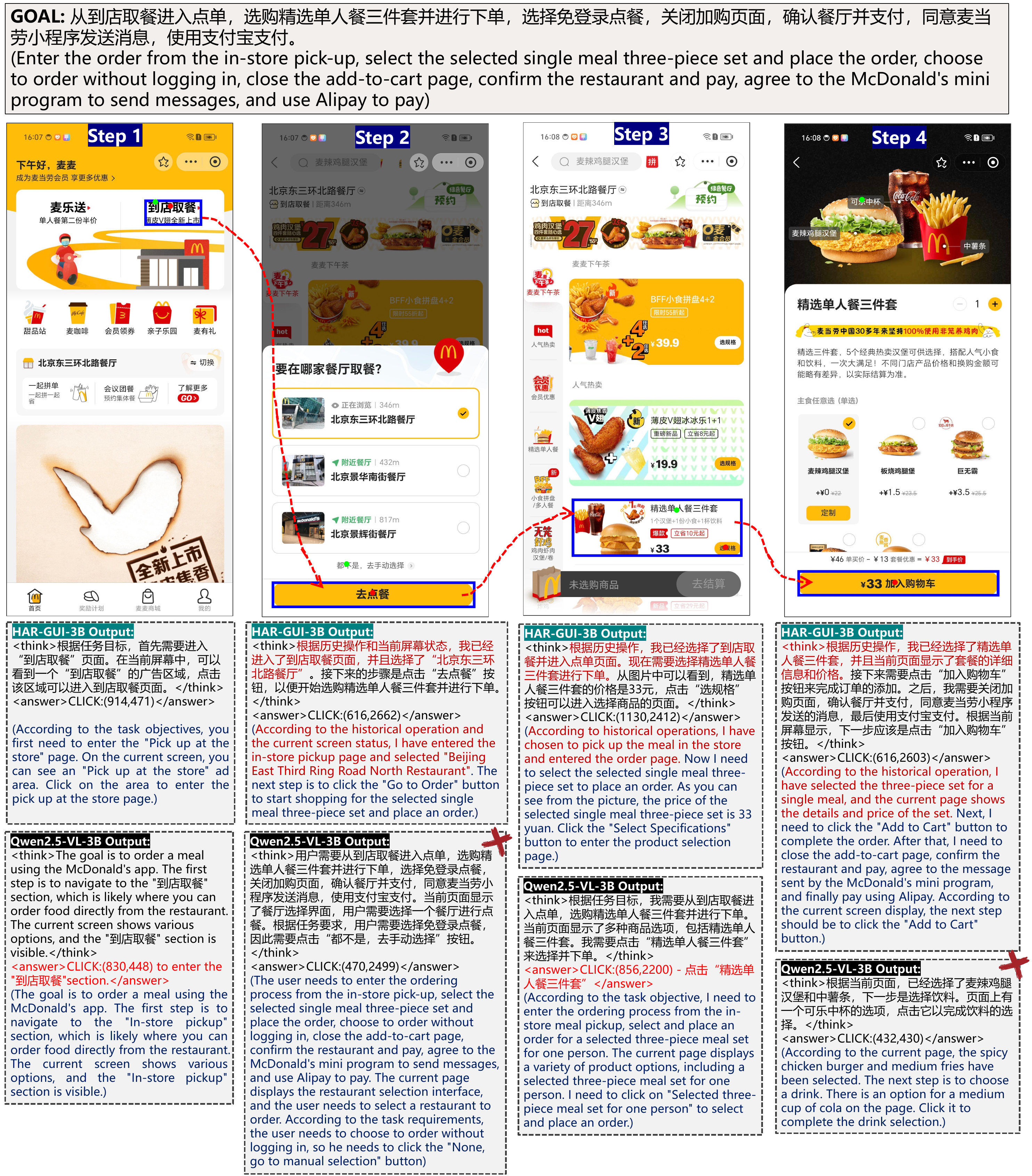}
    \caption{\textbf{Part A:} An episodic reasoning case from our in-house data in the \textbf{Takeout} scenario. The goal of episode is \textit{"Enter the order from the in-store pick-up, select the selected single meal three-piece set and place the order, choose to order without logging in, close the add-to-cart page, confirm the restaurant and pay, agree to the McDonald's mini program to send messages, and use Alipay to pay."}. The \textcolor{red}{\textbf{red dot}} indicates the prediction result of our \model, the \textcolor{green}{\textbf{green dot}} indicates the prediction result of Qwen2.5-VL-3B-Instruct, and the \textbf{bounding box} denotes the ground truth of the current step. It is evident that \model\ possesses stable short-term memory and effective screen details perception capability. On the contrary, the CoT of Qwen2.5-VL-3B-Instruct is history-agnostic, meaning that the chained long-horizon interaction is treated as a discrete screen understanding. Meanwhile, the decline of its instruction-following capability can also be observed.}
    \label{fig:ood-1-1}
\end{figure*}

\begin{figure*}
    \centering
    \includegraphics[width=1.0\linewidth]{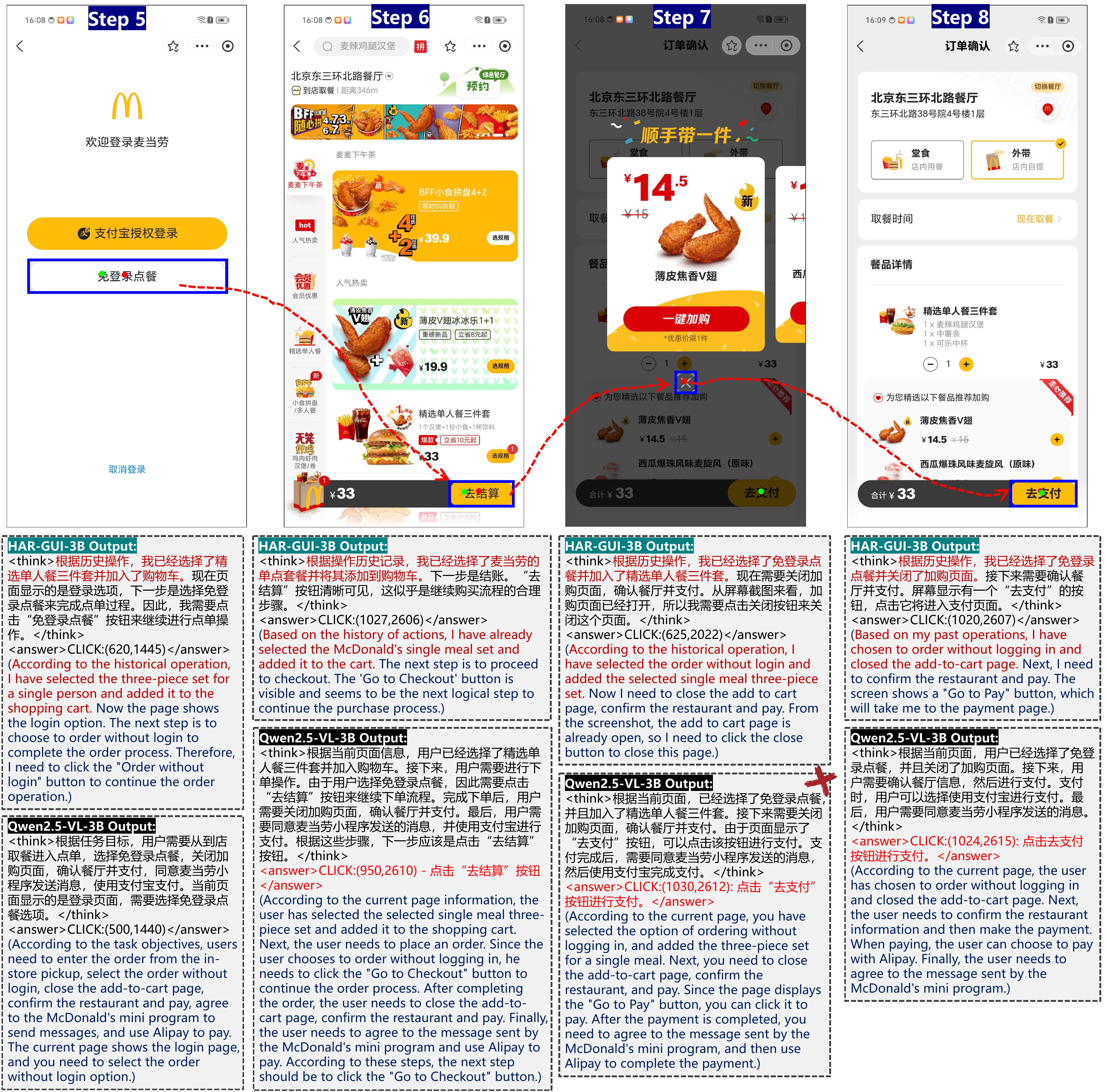}
        \caption{\textbf{Part B:} An episodic reasoning case from our in-house data in the \textbf{Takeout} scenario. The goal of episode is \textit{"Enter the order from the in-store pick-up, select the selected single meal three-piece set and place the order, choose to order without logging in, close the add-to-cart page, confirm the restaurant and pay, agree to the McDonald's mini program to send messages, and use Alipay to pay."}. The \textcolor{red}{\textbf{red dot}} indicates the prediction result of our \model, the \textcolor{green}{\textbf{green dot}} indicates the prediction result of Qwen2.5-VL-3B-Instruct, and the \textbf{bounding box} denotes the ground truth of the current step. It is evident that \model\ possesses stable short-term memory and effective screen details perception capability. On the contrary, the CoT of Qwen2.5-VL-3B-Instruct is history-agnostic, meaning that the chained GUI episode reasoning task is treated as an independent screen understanding task. Meanwhile, the decline of its instruction-following capability can also be observed.}
    \label{fig:ood-1-2}
\end{figure*}

\begin{figure*}
    \centering
    \includegraphics[width=1.0\linewidth]{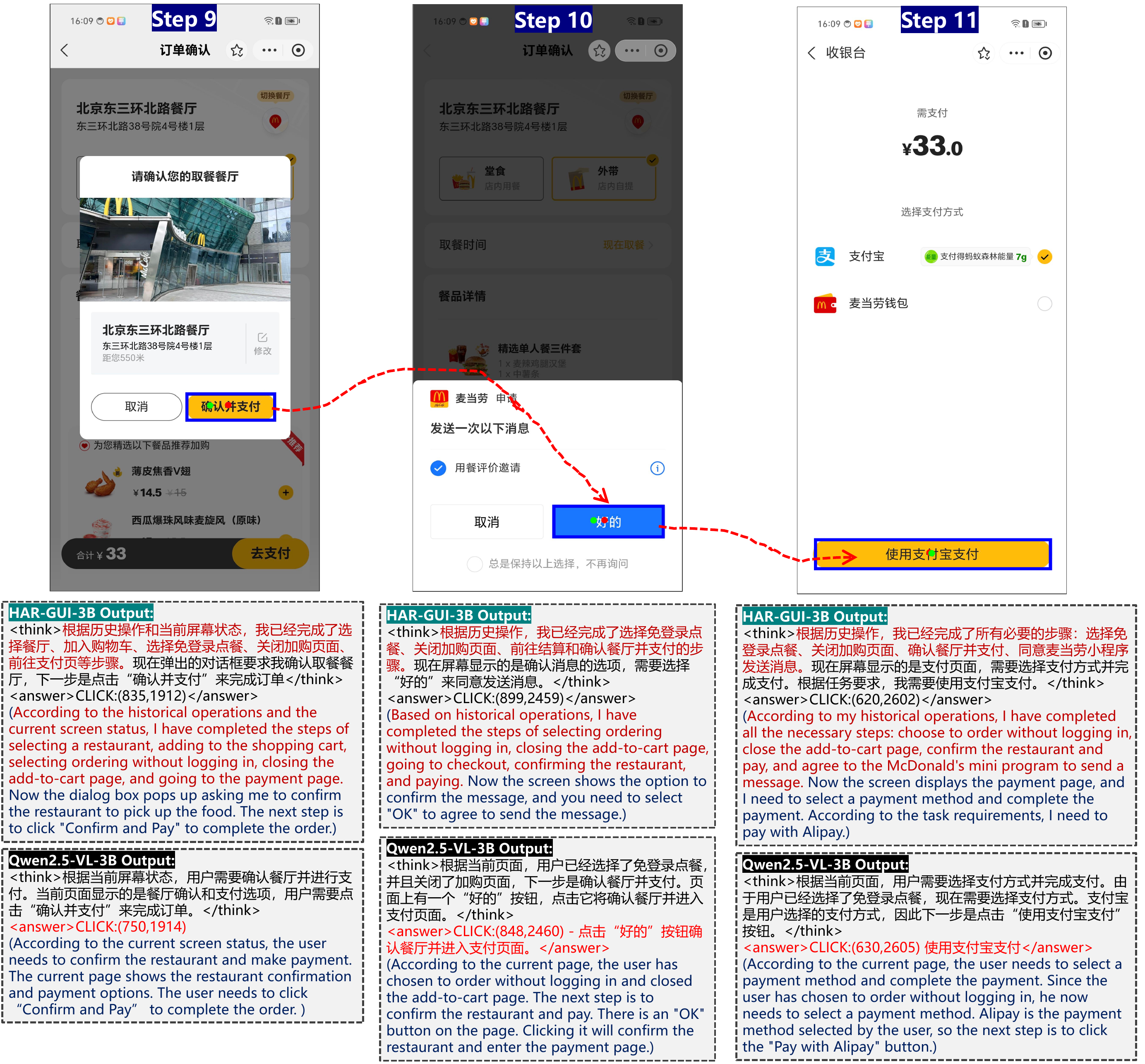}
        \caption{\textbf{Part C:} An episodic reasoning case from our in-house data in the \textbf{Takeout} scenario. The goal of episode is \textit{"Enter the order from the in-store pick-up, select the selected single meal three-piece set and place the order, choose to order without logging in, close the add-to-cart page, confirm the restaurant and pay, agree to the McDonald's mini program to send messages, and use Alipay to pay."}. The \textcolor{red}{\textbf{red dot}} indicates the prediction result of our \model, the \textcolor{green}{\textbf{green dot}} indicates the prediction result of Qwen2.5-VL-3B-Instruct, and the \textbf{bounding box} denotes the ground truth of the current step. It is evident that \model\ possesses stable short-term memory and effective screen details perception capability. On the contrary, the CoT of Qwen2.5-VL-3B-Instruct is history-agnostic, meaning that the chained GUI episode reasoning task is treated as an independent screen understanding task. Meanwhile, the decline of its instruction-following capability can also be observed.}
    \label{fig:ood-1-3}
\end{figure*}

As shown in Fig.\ref{fig:ood-1-1}, Fig.\ref{fig:ood-1-2} and Fig.\ref{fig:ood-1-3}. The goal of this episode is: \textit{"Enter the order from the in-store pick-up, select the selected single meal three-piece set and place the order, choose to order without logging in, close the add-to-cart page, confirm the restaurant and pay, agree to the McDonald's mini program to send messages, and use Alipay to pay."}, which is completed in 11 steps. We show the outputs of \model\ and Qwen2.5-VL-3B-Instruct. \textbf{It can be seen that our \model\ has a stable short-term memory and can effectively recognize historical interaction information to complete the prediction of the current time step} when performing the episodic reasoning task. Yet, Qwen2.5-VL-3B-Instruct treats the long-horizon interaction as discrete screen understanding and ignores chained interaction clues. At the same time, \textbf{we found that all baselines in Tab.\ref{tab:ood_performance} maintain the same history-agnostic reasoning mode as Qwen2.5-VL-3B-Instruct, that is, weak short-term memory.} On the other hand, we found that GUI-specific methods such as InfiGUI-R1-3B, GUI-R1-3B, UI-R1-3B, and UI-TARS-7B-DPO all have a \textbf{decline in instruction following capability} in our Chinese OOD scenario, while our \model\ does not have this situation. 

\subsubsection{The stable short-term memory of \model}
\begin{figure*}
    \centering
    \includegraphics[width=1.0\linewidth]{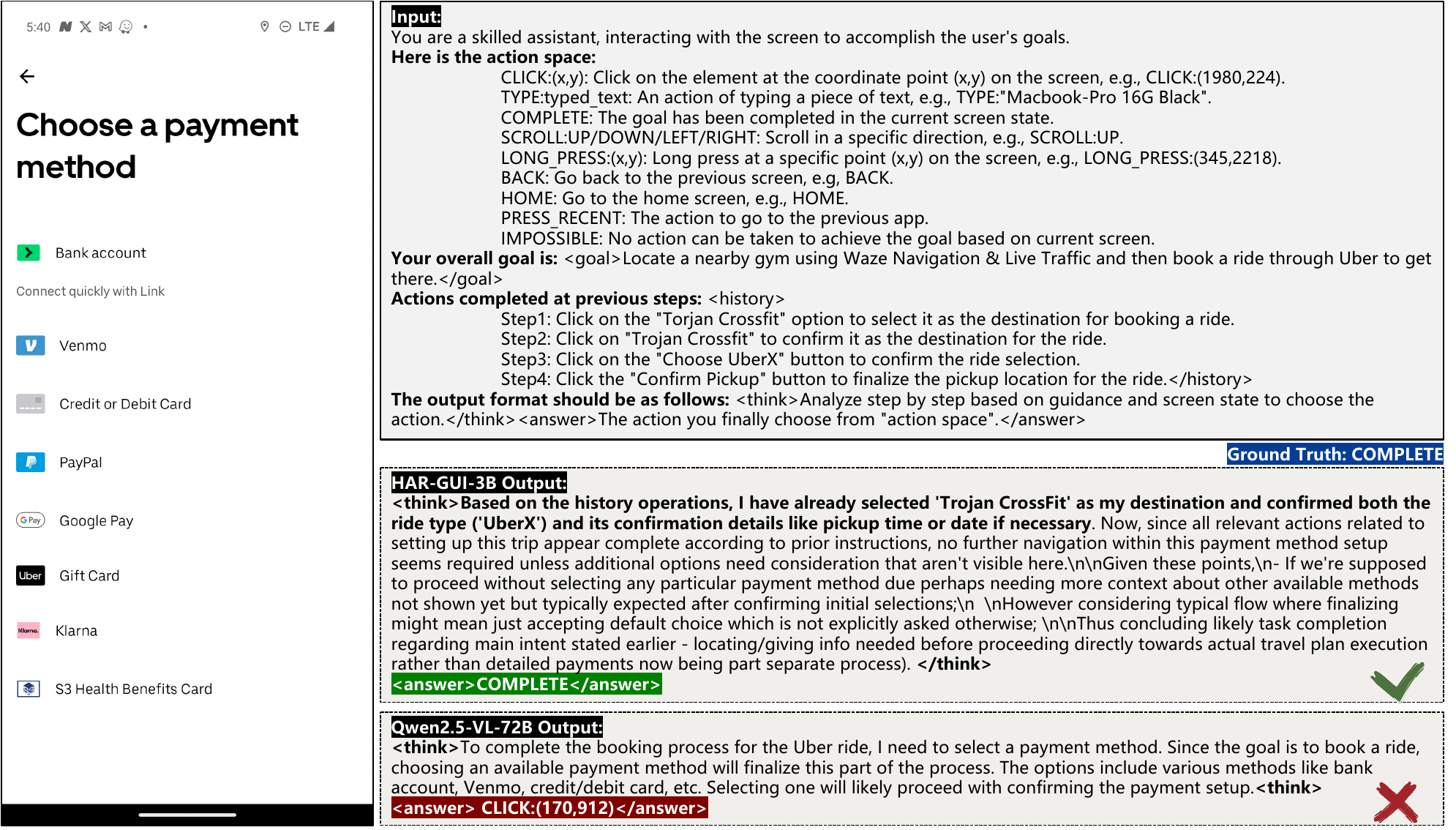}
    \caption{Comparison of Qwen2.5-VL-72B-Instruct and \model\ in processing GUI episodic reasoning \textbf{(Case-1)}. As shown in the figure, the history-conscious  reasoning mode of \model\ effectively captures the historical interaction context in the instruction. In contrast, the compared method \textbf{disregards sequential clues by treating the chained interactions as discrete screen understanding}, ultimately leading to incorrect predictions.}
    \label{fig:inference-demo1}
\end{figure*}

\begin{figure*}
    \centering
    \includegraphics[width=1.0\linewidth]{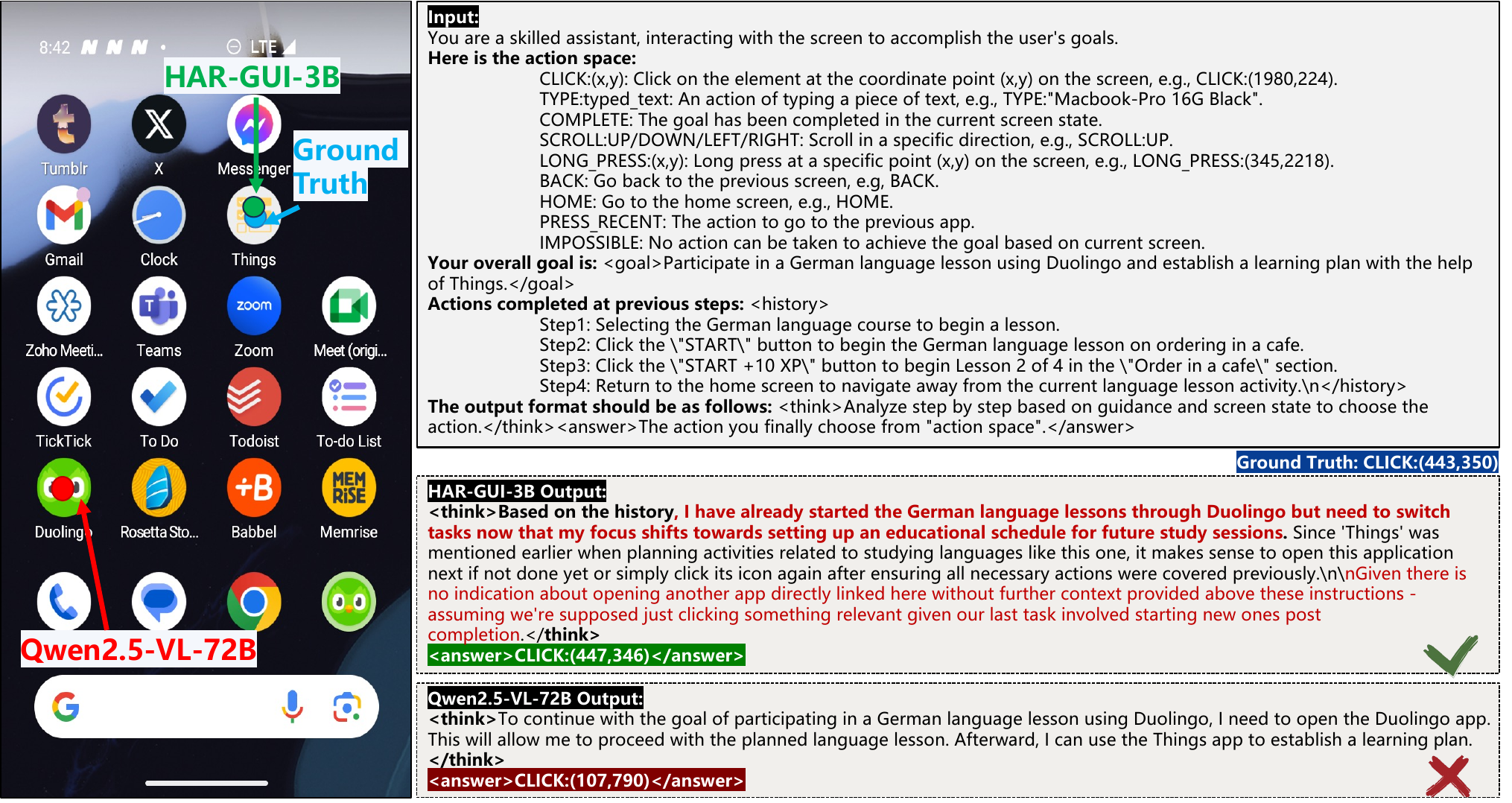}
    \caption{Comparison of Qwen2.5-VL-72B-Instruct and \model\ in processing episodic reasoning \textbf{(Case-2)}. As shown in the figure, the history-conscious  reasoning mode of \model\ effectively captures the historical interaction context in the instruction. In contrast, the compared method \textbf{disregards sequential clues by treating the chained interactions as discrete screen understanding}, ultimately leading to incorrect predictions.}
    \label{fig:inference-demo2}
\end{figure*}

\begin{figure*}
    \centering
    \includegraphics[width=1.0\linewidth]{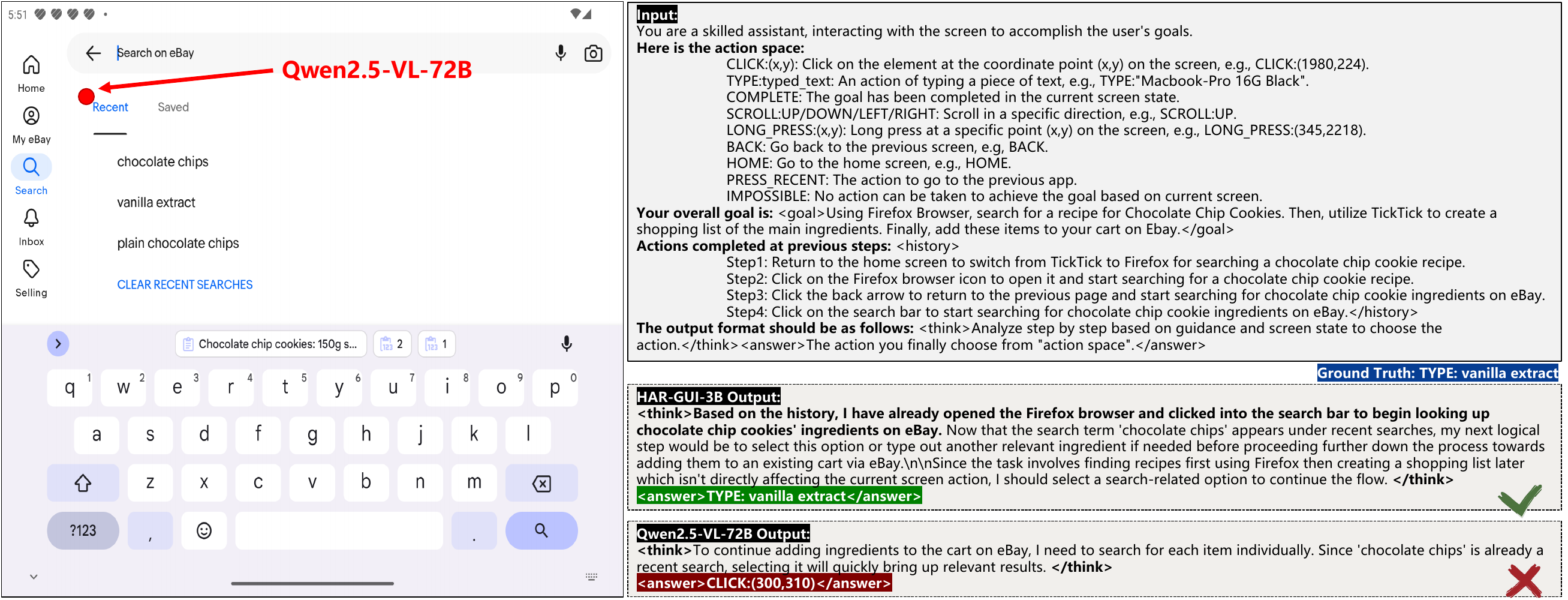}
    \caption{Comparison of Qwen2.5-VL-72B-Instruct and \model\ in processing episodic reasoning \textbf{(Case-3)}. \textbf{These results confirm the robust short-term memory and the capability to perceive screen details of \model}. Its rigorous logic and comprehensive reasoning mode empower it to effectively tackle long-horizon GUI tasks.}
    \label{fig:inference-demo3}
\end{figure*}

\begin{figure*}
    \centering
    \includegraphics[width=1.0\linewidth]{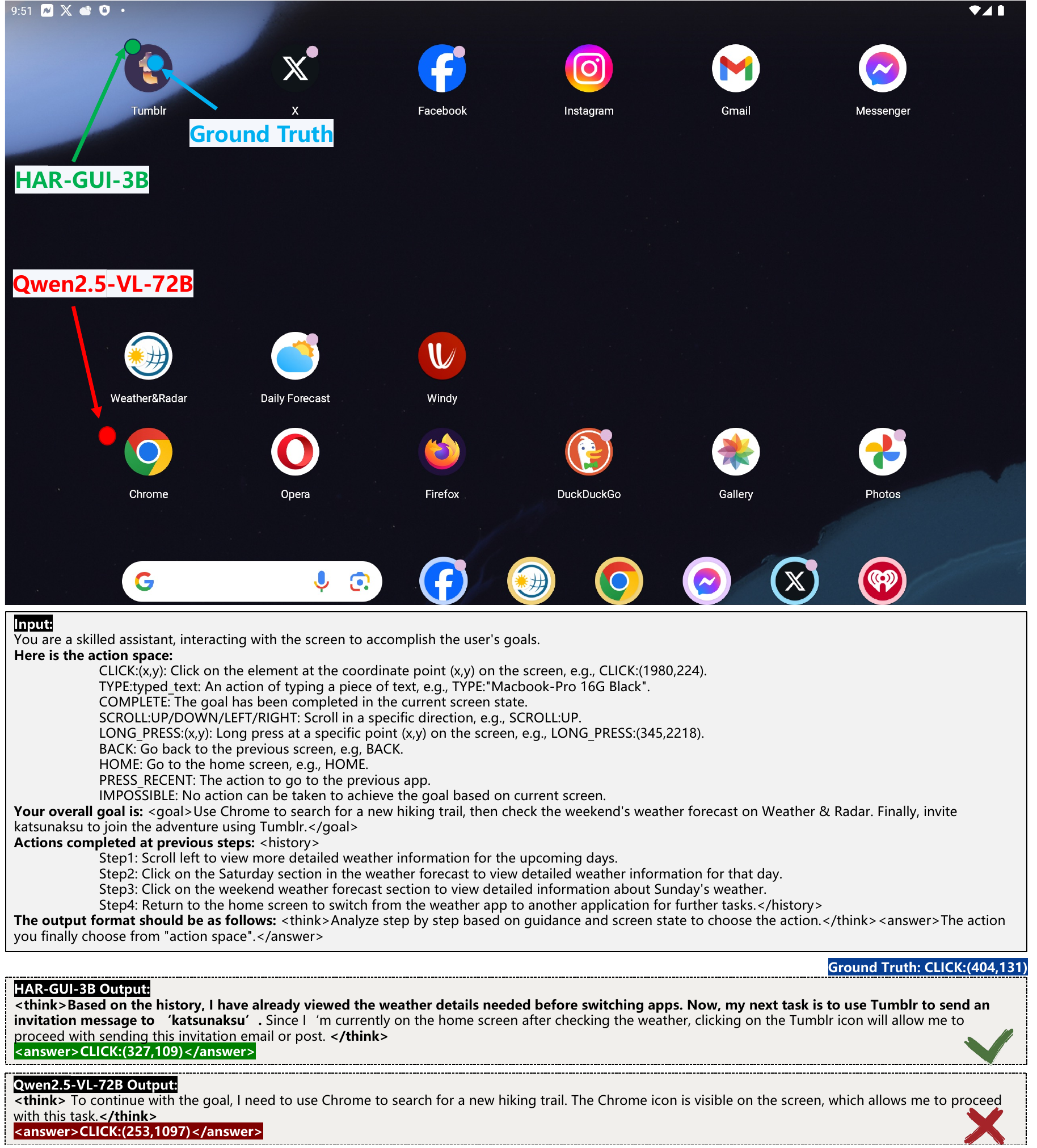}
    \caption{Comparison of Qwen2.5-VL-72B-Instruct and \model\ in processing episodic reasoning \textbf{(Case-4)}. As shown in the figure, the history-conscious  reasoning mode of \model\ effectively captures the historical interaction context in the instruction. In contrast, the compared method \textbf{disregards sequential clues by treating the chained interactions as discrete screen understanding}, ultimately leading to incorrect predictions.}
    \label{fig:inference-demo5}
\end{figure*}

In this part, we demonstrate the significance of the GUI agent's short-term memory in performing the GUI episode reasoning task by presenting execution results from the testset. Here, we select the powerful Qwen2.5-VL-72B-Instruct as the comparison method for problem analysis. As illustrated in Fig.\ref{fig:inference-demo1}, Fig.\ref{fig:inference-demo2}, Fig.\ref{fig:inference-demo3} and Fig.\ref{fig:inference-demo5}, \textbf{when Qwen2.5-VL-72B-Instruct processes the chained episodic reasoning, it also treats each execution step as a discrete screen understanding, neglecting the crucial sequential clues (historical interaction context). This history-agnostic reasoning mode is consistent with other advanced GUI agents we mentioned in the experimental section.} Clearly, overlooking the sequential clues hampers the GUI agent's capability to effectively conduct GUI episode reasoning task, with performance deteriorating further as the episode length increases. \textbf{In contrast, our \model\ maintains a stable short-term memory, enabling it to adeptly handle long sequence reasoning tasks through effective perception and analysis of historical interaction context.} Further, \model\ excels at both capturing screen visual clues and understanding GUI elements, such as pop-up window closing and UI elements grounding.
 
\begin{tcolorbox}[colback=blue!7, title = {The Template for Grounding Task}]
{\tt Locate the element on the screen with the function or description: [\textbf{DESCRIPTIONS}]. Keep the following output format: \{"point\_2d": [x,y], "label": description of the target element.\}}
\end{tcolorbox}

\begin{tcolorbox}[colback=blue!7, title = {The Template for Act2Sum Data Synthesis}]
{\tt Step-by-step GUI navigation task. Briefly summarize the current action.\\
{Action space: }[ACTION SPACE]\\
{Goal:} <goal>[GOAL]</goal>\\
{Current action:} <action>[ACTION]</action>\\
Output Format: <summary> One-sentence summary of the action based on the screen image.</summary>}
\end{tcolorbox}

\begin{tcolorbox}[colback=blue!7, title = {The Template for System-2 Episodic Reasoning (Inference-format Instruction)}]
{\tt You are a skilled assistant, interacting with the screen to accomplish the user's goals.\\
Here is the action space:
[ACTION SPACE]\\
Your overall goal is: <goal>[GOAL]</goal> \\
Actions completed at previous steps: <history> [PREVIOUSLY PERFORMED ACTIONS] </history>\\

The output format should be as follows:\\
<think>Analyze step by step based on guidance and screen state to choose the action.</think>\\
<answer>The action you finally choose from "action space".</answer>}
\end{tcolorbox}

\begin{tcolorbox}[colback=blue!7, title = {The Template for Tailored Guideline Synthesis}]
{\tt Below is a model's incorrect predictions when processing a GUI navigation task. Your goal is to identify the error's cause and provide helpful clues for correct future predictions.\\
Model Input: [\textbf{The Template for System-2 Episodic Reasoning}]\\
Model Output: <think> [ERROR THINK]</think> <answer>[ERROR ANSWER]</answer>\\
Ground Truth: [GROUND TRUTH ACTION]\\

Please keep the following output format:\\
<think> Please refer to the "Model Input" and compare the "Model Output" with the "Ground Truth" to analyze and output the reasons for the model prediction errors. </think>\\
<guidelines>Provide up to three helpful guidelines to guide the model to make correct prediction (keep it short and concise), formatted as a numbered list (\textit{e.g.}, 1.Guideline 1; 2.Guideline 2). Do not include content from the "Ground Truth".</guidelines>}
\end{tcolorbox}

\begin{tcolorbox}[colback=blue!7, title = {The Template for Error-Aware Cognitive Correction}]
{\tt You are a skilled assistant, interacting with the screen to accomplish the user's goals.\\
Here is the action space: [ACTION SPACE].\\
Your overall goal is: <goal>[GOAL]</goal> \\
Actions completed at previous steps: {<history>}[PREVIOUSLY PERFORMED ACTIONS] {</history>} \\
The output format should be as follows: <think> Analyze step by step based on guidance and screen state to choose the action. </think> <answer>The action you finally choose from "action space".</answer>\\

\textbf{However, you made an incorrect action prediction}: <think> [ERROR THINK]</think> <answer>[ERROR ANSWER]</answer>\\

\textbf{Here are some guidelines to help you select the right action}: {<guidelines> } [GUIDELINES] {</guidelines>}\\

The output format should be as follows:\\
{<statement>} Explain here why you made incorrect decision in the past and summarize your mistakes here. {</statement>} <think> Analyze step by step based on guidance and screen state to choose the action </think> <answer> The action you finally choose from "action space". </answer>}
\end{tcolorbox}

\begin{tcolorbox}[colback=blue!7, title = {The Template for System-1 Episodic Reasoning (Inference-format Instruction)}]
{\tt You are a skilled assistant, interacting with the screen to accomplish the user's goals.\\
Here is the action space:
[ACTION SPACE]\\
Your overall goal is: <goal>[GOAL]</goal> \\
Actions completed at previous steps: <history> [PREVIOUSLY PERFORMED ACTIONS] </history>\\

The output format should be as follows:\\
<answer>The action you finally choose from "action space".</answer>}
\end{tcolorbox}

\begin{tcolorbox}[colback=blue!7, title = {The Template for Model-based Memory-Augmented Reward (MAR)}]
{\tt The following is the output of an agent performing an GUI episodic reasoning task in an environment, representing the reflection and thinking process of the agent during task execution. \\

Below are the actions that the agent has previously completed:
<history>[PREVIOUSLY PERFORMED ACTIONS] </history>\\

Below is the agent's output:
[OUTPUT]\\

***Your task***\newline
Since historical interaction records (what the agent has done before) are the key clues for the agent to make correct sequence predictions, you need to determine whether the content of the "<think> ... </think>" tag in the agent's output directly contains the previously completed interaction information contained in the "<history> ... </history>" tag. 
\newline
\newline
If it has been considered, output "yes" and the reason; otherwise, output "no".}
\end{tcolorbox}





\end{document}